\pdfoutput=1

\documentclass[11pt]{article}

\usepackage{EMNLP2023}

\usepackage{times}
\usepackage{latexsym}
\usepackage{multirow}
\usepackage{graphicx}
\usepackage{amsmath,amsthm}

\usepackage[T1]{fontenc}

\usepackage[utf8]{inputenc}

\usepackage{microtype}

\usepackage{inconsolata}

\title{LLM-Adapters: An Adapter Family for Parameter-Efficient Fine-Tuning of Large Language Models}

\author{
    ~~Zhiqiang Hu$^{1}$
    ~~Lei Wang$^{2}$ $\thanks{~~Corresponding author.}$
    ~~Yihuai Lan
    ~~Wanyu Xu$^{4}$  
    ~~Ee-Peng Lim$^{2}$  \\ \bf
    ~~Lidong Bing$^{3}$  
    ~~Xing Xu$^{5}$   
    ~~Soujanya Poria$^{1}$  
    ~~Roy Ka-Wei Lee$^{1}$\\ 
    $^{1}$Singapore University of Technology and Design\\
    $^{2}$Singapore Management University \\
    $^{3}$DAMO Academy, Alibaba Group \\
    $^{4}$Southwest Jiaotong University \\
    $^{5}$University of Electronic Science and Technology of China \\
}

\begin{document}
\maketitle
\begin{abstract}

The success of large language models (LLMs), like GPT-4 and ChatGPT, has led to the development of numerous cost-effective and accessible alternatives that are created by finetuning open-access LLMs with task-specific data (e.g., ChatDoctor) or instruction data (e.g., Alpaca). Among the various fine-tuning methods, adapter-based parameter-efficient fine-tuning (PEFT) is undoubtedly one of the most attractive topics, as it only requires fine-tuning a few external parameters instead of the entire LLMs while achieving comparable or even better performance. To enable further research on PEFT methods of LLMs, this paper presents LLM-Adapters, an easy-to-use framework that integrates various adapters into LLMs and can execute these adapter-based PEFT methods of LLMs for different tasks. The framework includes state-of-the-art open-access LLMs such as LLaMA, BLOOM, and GPT-J, as well as widely used adapters such as Series adapters, Parallel adapter, Prompt-based learning and Reparametrization-based methods. Moreover, we conduct extensive empirical studies on the impact of adapter types, placement locations, and hyper-parameters to the best design for each adapter-based methods. We evaluate the effectiveness of the adapters on fourteen datasets from two different reasoning tasks, Arithmetic Reasoning and Commonsense Reasoning. The results demonstrate that using adapter-based PEFT in smaller-scale LLMs (7B) with few extra trainable parameters yields comparable, and in some cases superior, performance to powerful LLMs (175B) in zero-shot inference on both reasoning tasks.
\end{abstract}

\section{Introduction}

Large language models (LLMs), such as ChatGPT~\cite{openai-chatgpt-2022} and GPT-4~\cite{openai-gpt4-2023}, have demonstrated unprecedented performance across various natural language processing (NLP) tasks~\cite{qin-chatgpt-2023} and multi-modal tasks~\cite{hugginggpt}. These LLMs often possess sizes exceeding hundreds of billions of parameters and are closed-source. Consequently, this has spurred the development of accessible and cost-effective alternatives such as LLaMA~\cite{llama}.
These alternatives involve fine-tuning open-source LLMs utilizing either task-specific data (e.g., ChatDoctor~\cite{yunxiang2023chatdoctor}) or instructional data (e.g., Alpaca~\cite{alpaca}). However, full-model fine-tuning (FFT) is computationally and storage-intensive, thereby presenting significant challenges in practical implementation.


Prior to the emergence of FFT of LLMs (e.g., LLaMA), a compelling solution called parameter-efficient fine-tuning (PEFT)~\cite{adapters} has been proposed in the NLP field, specifically for pre-trained models (e.g., BERT~\cite{devlin2018bert}), offering a promising approach for efficiently fine-tuning LLMs. 
The advantage of PEFT lies in its ability to fine-tune only a small set of external parameters rather than the entire backbone model while still achieving comparable or even superior performance~\cite{peft}.
Moreover, PEFT can effectively mitigate catastrophic forgetting in comparison to FFT~\cite{adamix}.
As shown in Table~\ref{tab:peft_category}, the advantage of PEFT has resulted in the developing of diverse PEFT modules, encompassing series adapters~\cite{adapters, adamix, sparseadapter, lets}, parallel adapters~\cite{parallel_adapter}, reparameterization-based methods~\cite{lora, krona}, and prompt-based learning methods~\cite{prompt_tuning, prefix}.

By incorporating these PEFT modules into backbone models (i.e., LLMs), we can capitalize on the remarkable capabilities of backbone models without requiring extensive computational resources. 
This opens up opportunities for a broader range of applications, enabling even those with limited access to high-performance computing to harness the power of LLMs in their specific tasks.
Despite the success of PEFT for pre-trained models, it remains unclear which PEFT module, in combination with which layer and hyperparameter configuration, is most suitable for a given task or dataset when meeting LLMs (e.g., LLaMA~\cite{llama}). Therefore, further investigation is needed to determine the optimal PEFT setup that maximizes performance across different tasks and datasets.

Motivated by this, in this paper, we conduct a comprehensive empirical study of PEFT of three representative open-source LLMs, including BLOOM~\cite{bloom}, GPT-J~\cite{gpt-j}, and LLaMA~\cite{llama}.
Specifically, we undertake an empirical study to address the following three research questions: ($i$) What is the optimal placement and configuration of different PEFT methods? ($ii$) How's the performance of different adapters across downstream tasks? And ($iii$) What are the differences in performance between in-distribution (ID) and out-of-distribution (OOD) scenarios for PEFT methods? 
The findings of our study are as follows: 
\begin{enumerate}
    \item \textbf{The optimal placement for the series adapter, parallel adapter, and LoRA is after the MLP layers, parallel with the MLP layers, and located after both the Attention layers and MLP layers simultaneously, respectively};
    \vspace{-5pt}
    \item \textbf{Smaller language models with the PEFT approach can attain competitive or superior performance on specific tasks compared to larger language models.
    For instance, LLaMA-13B with LoRA can outperform GPT-3.5 (>175B) on MultiArith, AddSub, and SingleEq
    }; 
    \vspace{-5pt}
    \item \textbf{The ID fine-tuned LLaMA-13B with adapters outperforms ChatGPT on commonsense reasoning tasks indicating that smaller language models have the potential to outperform larger language models on specific tasks with ID fine-tunig data.} 
\end{enumerate}

Our contributions can be summarized as follows:
\begin{itemize}
    \item We conduct a comprehensive empirical study of various PEFT methods applied in different open-source LLMs. 
    \vspace{-5pt}
    \item To facilitate our empirical study, we construct two high-quality training datasets to enhance PEFT performance in math reasoning and commonsense reasoning tasks. 
    \vspace{-5pt}
    \item We develop a user-friendly framework, LLM-Adapter, seamlessly integrates diverse adapters into LLMs, empowering researchers to implement adapter-based PEFT methods for a wide range of tasks. 
    \vspace{-5pt}
    \item We conduct extensive experiments to answer the three research questions to serve as inspiration for future research. 
\end{itemize}


\section{PEFT Overview}

\begin{figure*}[t] 
	\centering
	\includegraphics[scale = 0.12]{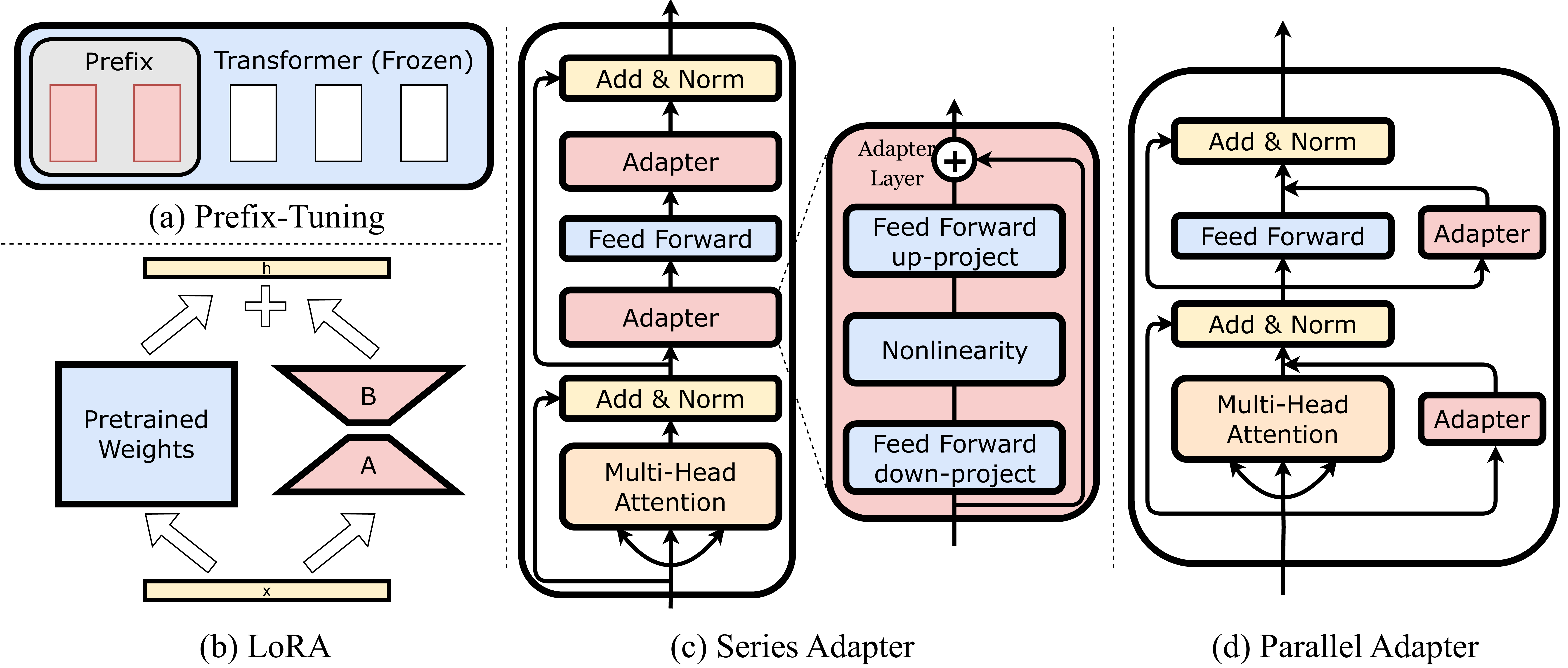} 
	\vspace{-2pt}
	\caption{A detailed illustration of the model architectures of three different adapters: (a) Prefix-Tuning, (b) LoRA, (c) Series Adapter, and (d) Parallel Adapter.}
        \label{fig:peft_overview}
	\vspace{-8pt}
\end{figure*}

\begin{table}[tb]
\centering\small
\vspace{-5pt}
\setlength{\tabcolsep}{1pt}
\resizebox{0.48\textwidth}{!}{
\begin{tabular}
{l|cccc}
\hline
Method & Prompt & Repara & Series & Parallel  \\\hline
Prompt Tuning \cite{prompt_tuning}      &$\surd$ & & & \\
Prefix-Tuning  \cite{prefix}    & $\surd$& & & \\
Spot   \cite{spot}            & $\surd$& & & \\
IPT    \cite{ipt}        &$\surd$ & & & \\
LoRA     \cite{lora}       & &$\surd$ & & \\
KronA   \cite{krona}       & &$\surd$ & & \\
Adapters \cite{adapters}  & & &$\surd$ &        \\
AdaMix \cite{adamix}  & & &$\surd$ & \\
SparseAdapter \cite{sparseadapter}       & & &$\surd$ & \\
LeTS    \cite{lets}      & & &$\surd$ & \\
Parallel Adapter \cite{parallel_adapter} & & & & $\surd$ \\
MAM Adapter  \cite{mam}    &$\surd$ &$\surd$ &$\surd$ & \\
UniPELT   \cite{unipelt}      &$\surd$ & $\surd$ &$\surd$ & \\
Compacter  \cite{compacter}     & &$\surd$ &$\surd$ & \\
S4-model  \cite{s4_model}     &$\surd$ &$\surd$ & & \\
\hline
\end{tabular}
}
\vspace{-2pt}
\caption{The PEFT methods are categorized based on the four common basic methods. "Prompt" represents prompt-based learning methods, "Repara" denotes reparametrization-based methods, "Series" is Series Adapter, while "Parallel" represents Parallel Adapter.
}
\label{tab:peft_category}
\vspace{-8pt}
\end{table}

In this section, we provide a brief overview of four parameter-efficient fine-tuning (PEFT) methods: prompt-based learning, reparametrization-based methods, series adapters, and parallel adapters. \cite{prefix,lora,adapters,parallel_adapter}

\paragraph{Prompt-based learning.} As shown in Figure~\ref{fig:peft_overview}(a), prompt-based learning transforms the discrete optimization problem of finding the optimal hard prompt into a continuous (soft) prompt. To achieve this, \citet{prompt_tuning} proposed the concept of prompt tuning, where a trainable tensor is added as a prefix to the input embeddings. Another approach called Prefix Tuning\cite{prefix} independently explored the addition of soft prompts to the hidden states of all layers. Intrinsic Prompt Tuning~\cite{ipt} employs an autoencoder to compress and decompress the soft prompt.
We take learnable vectors incorporated into the attention layer as an example of prompt-based learning, which can be formulated as follows:
\begin{align}
    & H_{o}^{} = \text{Attn}(H_{i}W_{Q},[P_{K};H_{i}W_{K}], [P_{V};H_{i}W_{V}]),
\end{align}
where $H_i\in\mathrm{R}^{T\times d}$ and $H_o\in\mathrm{R}^{T\times d}$ are the input and output of the attention layer respectively. Note that $T$ is the maximum input length and $d$ is the vector dimension. $P_{K}\in\mathrm{R}^{L\times d}$ and $P_{V}\in\mathrm{R}^{L\times d}$ are the learnable vectors for PEFT. 
$L$ is the number of learnable tokens, which is discussed in the experiment section in detail.
$Q, K, V$ denote the query, key, value vectors of th attention module, respectively.

\paragraph{Reparametrization-based method.} This type of methods aim to transform network weights using a low-rank technique. This approach effectively reduces the number of trainable parameters while preserving the ability to handle high-dimensional matrices. Intrinsic SAID~\cite{intrinsic_said} investigates the intrinsic dimensionality of fine-tuning within a low-rank subspace. LoRA~\cite{lora} introduces a simple approach to update the parameters of a weight matrix by decomposing it into a product of two low-rank matrices. KronA~\cite{krona} improves upon the matrix factorization aspect of LoRA by utilizing the Kronecker product in its technique.
We take LoRA as an example of  Reparametrization-based learning, which can be formulated below:
\begin{align}
H_{o}=H_{i}W_{0}+H_{i}\Delta W=H_{i}W_{0}+ H_{i}BA,
\end{align}
where $W_0\in\mathrm{R}^{d\times d}$ can be any pre-trained weight matrix, including weights in the MLP or Attention layer. $B\in\mathrm{R}^{r\times d}$ and $A\in\mathrm{R}^{r\times d}$ are lower-rank matrix intended for covering $\Delta W$. $r \ll d$ is an important hyper-parameter for LoRA.

\paragraph{Series Adapter.} Series adapters involve incorporating additional learnable modules in a sequential manner within a specific sublayer. In their study, \citet{adapters} proposed integrating fully-connected networks after the attention and FFN layers in the Transformer model~\cite{vaswani2017attention}. Another finding by \citet{madx} revealed that achieving comparable performance is possible by inserting the adapter solely after the self-attention layer, instead of using two adapters per transformer block. AdaMix~\citep{adamix} introduces a method that utilizes multiple series adapters in a mixture-of-experts (MoE) fashion. Compacter~\citep{compacter} utilizes the Kronecker product, low-rank matrices, and parameter sharing across layers to generate adapter weights. This technique aims to reduce the computational complexity associated with the adapters while maintaining their performance. 
Series Adapter can be formulated as follows:
\begin{equation}
     H_{o} \gets  H_{o}+f( H_{o}W_{down})W_{up},
\end{equation}
where the output $H_{o}$ of a specific layer, such as the MLP layer, is first down-projected by $W_{down}\in\mathrm{R}^{d\times r}$ to a lower dimension $r$, and then up-projected back by $W_{up}\in\mathrm{R}^{r\times d}$ to the original dimension $d$. $f$ is a non-linear function. We discuss the choice of $r$ in the experiment Section.

\paragraph{Parallel Adapter.} Parallel adapters \cite{parallel_adapter}  aim to incorporate additional learnable modules in parallel with distinct sublayers within the backbone model. 
The parallel adapter can be formulated below:
\begin{equation}
     H_{o} \gets  H_{o}+f( H_{i}W_{down})W_{up},
\end{equation}
where $H_{i}$ ($H_{o}$) is the input (output) of a specific layer.
Expanding on this concept, the Multi-head Parallel Adapter takes it a step further by using parallel adapters to modify the outputs of head attention. On the other hand, the Scaled Parallel Adapter is a variant that applies the composition and insertion format of LoRA~\cite{lora} to adapters. Another approach, called Ladder Side-Tuning~\cite{ladder_side_tuning}, involves training a lightweight ladder side network. This network accepts intermediate activations from the backbone networks through shortcut connections (ladders).

\section{Experiment Setup}
\subsection{Benchmarks}
We conduct extensive empirical studies on fourteen benchmark datasets from two categories of reasoning problems: 
\textbf{Arithmetic Reasoning:} (1) the {GSM8K}~\citep{gsm8k} dataset consists of high quality linguistically diverse grade school math word problems created by human problem writers, (2) the {SVAMP}~\citep{svamp} benchmark consists of one-unknown arithmetic word problems for up-to-4 grade level students by making simple changes to a set of problems from another existing dataset, (3) the {MultiArith}~\citep{mutli_arith} dataset of math word problems requiring multiple reasoning steps and operations, (4) the {AddSub}~\cite{addsub} dataset of addition and subtraction arithmetic word problems, (5) the {AQuA}~\citep{aqua} dataset of algebraic word problems with natural language rationales, and (6) the {SingleEq}~\citep{singleeq} dataset of grade-school algebra word problems that map to single equations with varying length; 
\textbf{Commonsense Reasoning:} (1) the BoolQ \cite{boolq} dataset is a question-answering dataset for yes/no questions containing 15942 examples. These questions are naturally occurring and generated in unprompted and unconstrained settings, (2) the PIQA \cite{piqa} dataset of questions with two solutions requiring physical commonsense to answer, (3) the SIQA \cite{siqa} focuses on reasoning about people's actions and their social implications, (4) the HellaSwag dataset of commonsense NLI questions including a context and several endings which complete the context, (5) the WinoGrande \cite{winogrande} dataset is formulated as a fill-in-a-blank task with binary options, and the goal is to choose the right option for a given sentence which requires commonsense reasoning, (6) the ARC-c and (7) the ARC-e are the Challenge Set and Easy Set of ARC \cite{arc} dataset of genuine grade-school level, multiple-choice science questions, and (8) the OBQA dataset contains questions requiring multi-step reasoning, use of additional common and commonsense knowledge, and rich text comprehension. Table \ref{tab:dataset_description} shows the dataset statistics.

\begin{table}[tb]
\centering\small
\vspace{-5pt}
\setlength{\tabcolsep}{3pt}
\begin{tabular}
{lcccr}
\hline
Dataset & Domain & \# train & \# test &Answer  \\\hline
{MultiArith} & Math   & -    &600      &Number \\
{AddSub}     & Math   & -    &395      &Number \\
{GSM8K}      & Math   & 8.8K &1,319    &Number \\
{AQuA}       & Math   & 100K &254      &Option \\
{SingleEq}   & Math   & -    &508      &Number \\
{SVAMP}      & Math   & -    &1,000    &Number \\
{BoolQ}      & CS     & 9.4K &3,270    &Yes/No \\
{PIQA}       & CS     & 16.1K&1,830    &Option \\
{SIQA}       & CS     & 33.4K&1,954    &Option \\
{HellaSwag}  & CS     & 39.9K&10,042   &Option \\
{WinoGrande} & CS     & 63.2K&1,267    &Option \\
{ARC-e}      & CS     & 1.1K &2,376    &Option \\
{ARC-c}      & CS     & 2.3K &1,172    &Option \\
{OBQA}       & CS     & 5.0K &500      &Option \\
\hline
\end{tabular}
\vspace{-2pt}
\caption{Details of datasets being evaluated. Math: arithmetic reasoning. CS: commonsense reasoning.}
\label{tab:dataset_description}
\vspace{-8pt}
\end{table}

\subsection{Fine-tuning Data Collection}
In order to perform fine-tuning on adapters, we acquire two high-quality training datasets specifically designed for math reasoning and commonsense reasoning. Table \ref{tab:dataset_description} reveals that only GSM8K and AQuA datasets provide training sets for arithmetic reasoning. To enhance the diversity of our data, we incorporate the training sets from GSM8K, MAWPS, MAWPS-single \cite{mawps}, and select 1000 examples from AQuA for the purpose of collecting the fine-tuning data. However, it is worth noting that the chosen datasets solely offer equations and corresponding answers. In order to augment the reasoning capabilities of our model, particularly in terms of providing step-by-step rationales, we leverage ChatGPT as the teacher model. By utilizing zero-shot chain-of-thought prompts, ChatGPT generates reasoning steps. We have included the specific prompt templates used to collect the math reasoning dataset in Appendix \ref{sec:math_templates}. To ensure the quality of the data, we eliminate samples that contain incorrect answers. As a result, we obtain a set of 10K math reasoning samples, referred to as Math10K, which we consider for further analysis and fine-tuning. 

To facilitate fine-tuning in the domain of commonsense reasoning, we construct fine-tuning data by formatting the training sets from BoolQ, PIQA, SIQA, HellaSwag, WinoGrande, ARC-e, ARC-c, and OBQA with pre-defined templates. As each dataset in the commonsense reasoning domain entails distinct tasks, we adopt a structured template by initially describing the task's goal, followed by the corresponding content and answer. The template utilized for creating the fine-tuning data can be found in \ref{sec:commonsense_templates}. Upon completion of this process, we obtain a collection of 170K commonsense reasoning samples, which we refer to as Commonsense170K. These datasets will be made publicly available to encourage further research and exploration in this area. 

\subsection{Implementations}
To facilitate the seamless utilization of PEFT methods in both research and practical applications, we have developed a user-friendly framework, LLM-Adapter. LLM-Adapters seamlessly integrates diverse adapters into LLMs, empowering researchers to implement adapter-based PEFT methods for a wide range of tasks. We utilize LLaMA (7B, 13B) \cite{llama}, BLOOMz (7B) \cite{bloom}, and GPT-J (6B) \cite{gpt-j} as the base models for our experiments. As for the four categories of PEFT methods, we select Prefix-Tuning \cite{prefix}, Series Adapter \cite{adapters}, LoRA \cite{lora}, and Parallel adapter \cite{parallel_adapter} as representative candidates to examine their efficacy. For consistency across all fine-tuning experiments, we maintain a batch size of 16. The learning rate for Prefix-Tuning is set to 3e-2, while the rest of the methods adopt a learning rate of 3e-4. Each of the PEFT methods is fine-tuned for three epochs on the fine-tuning datasets. It is important to note that we fine-tune a single model for either the math or commonsense reasoning task, and subsequently evaluate its performance across all corresponding datasets.

\section{Experiment Results}
\subsection{Placement and Configuration}
To address the research question, \textit{``What is the optimal placement and configuration for various types of adapters?''}, we employ LLaMA-7B as the base model to assess different adapter settings within the context of the math reasoning task. Our empirical study begins by determining the most effective placement for the Series Adapter, Parallel Adapter, and LoRA. Prefix-Tuning is excluded from this analysis since its placement is predetermined. For the Series Adapter, we explore its placement options after the multi-head attention layers, MLP layers, or both of them. As for the Parallel Adapter and LoRA, we integrate them into the multi-head attention layers, MLP layers, or both of them, in order to assess their respective performances. The detailed results on each dataset are shown in Appendix \ref{placement_analysis}.  Figure \ref{fig:placement} shows the average accuracy on math reasoning datasets. We can observe that for the Series Adapter, the best position is to place it after the MLP layers, achieving an average accuracy of $59.5\%$ on the math reasoning datasets. As for the Parallel Adapter, when we place it within the MLP layers, it achieves the best performance of $61.7\%$. Regarding LoRA, we need to insert it simultaneously into both the Multi-head Attention layers and MLP layers to achieve the best performance of $60\%$. 

\begin{figure}[t] 
	\centering
	\includegraphics[scale = 0.35]{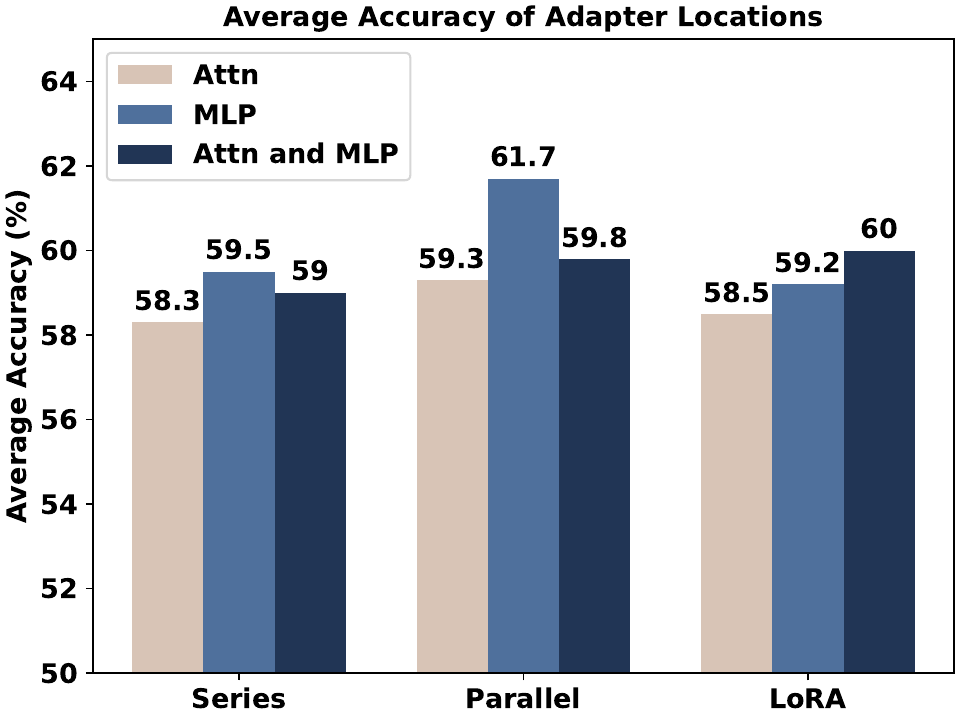} 
	\vspace{-5pt}
	\caption{The average accuracy of different adapter locations on math reasoning datasets.}
        \label{fig:placement}
\end{figure}

\begin{figure}[t] 
    \centering
    \begin{tabular}{cc}
	\includegraphics[scale = 0.22]{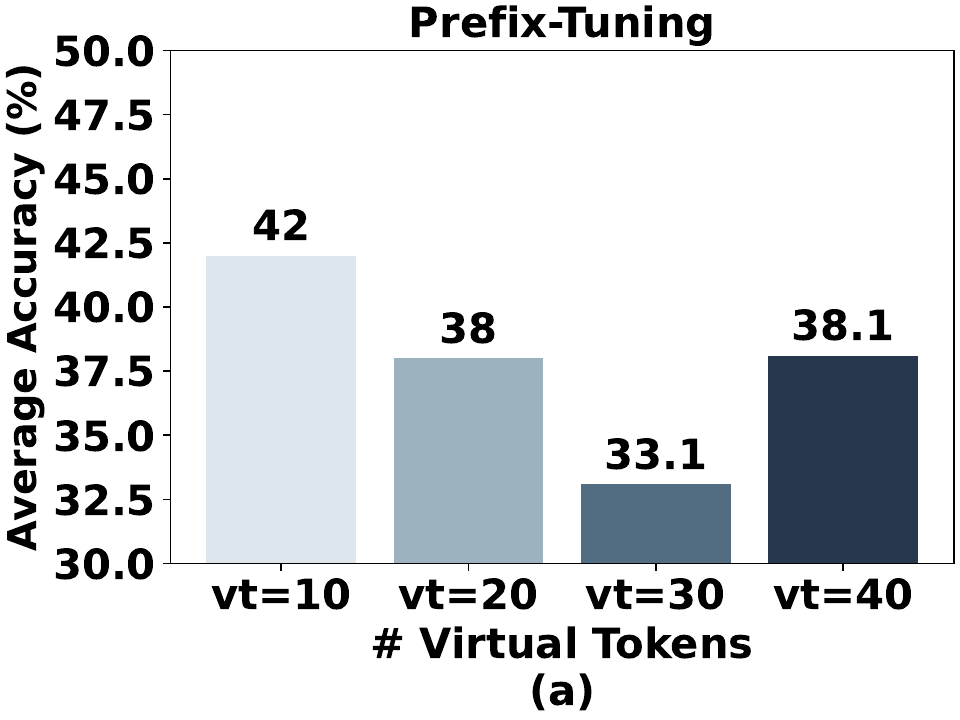} &
        \includegraphics[scale = 0.22]{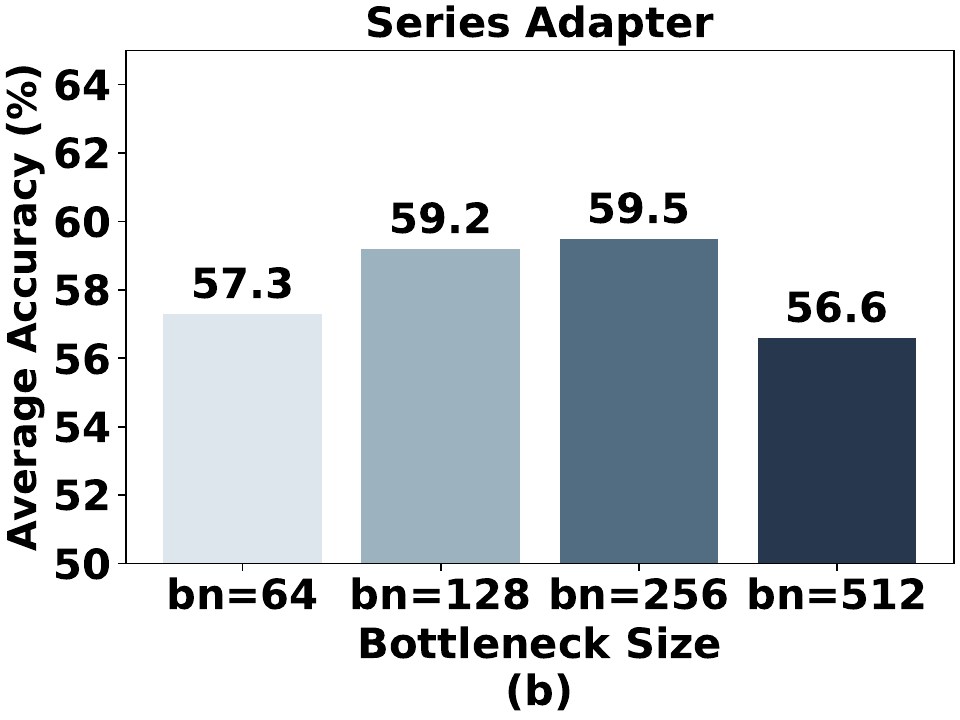} \\
        \includegraphics[scale = 0.22]{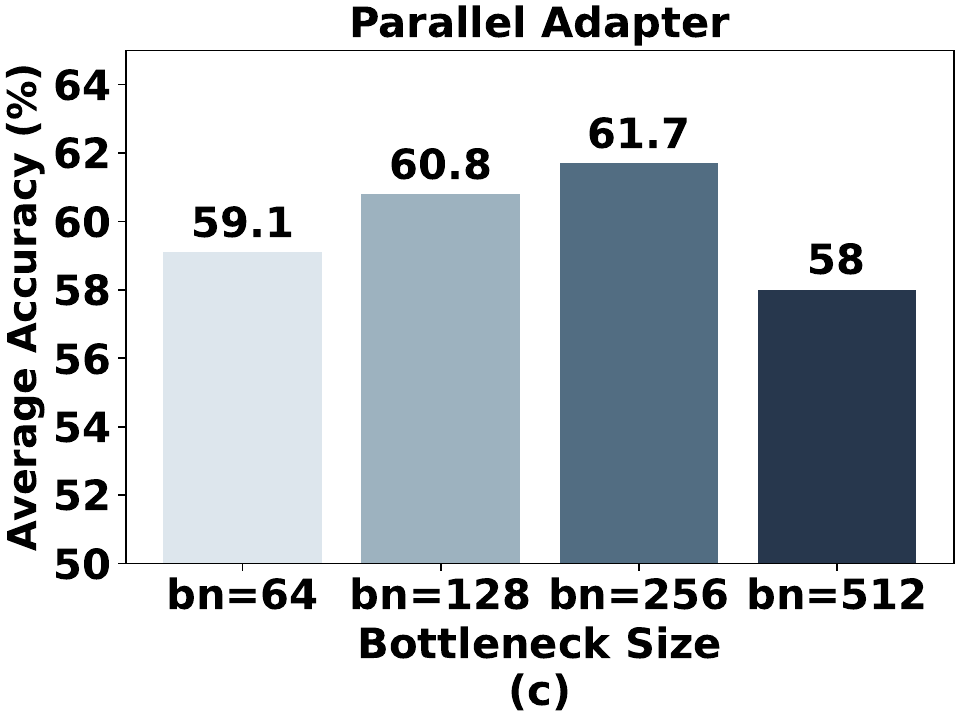} &
        \includegraphics[scale = 0.22]{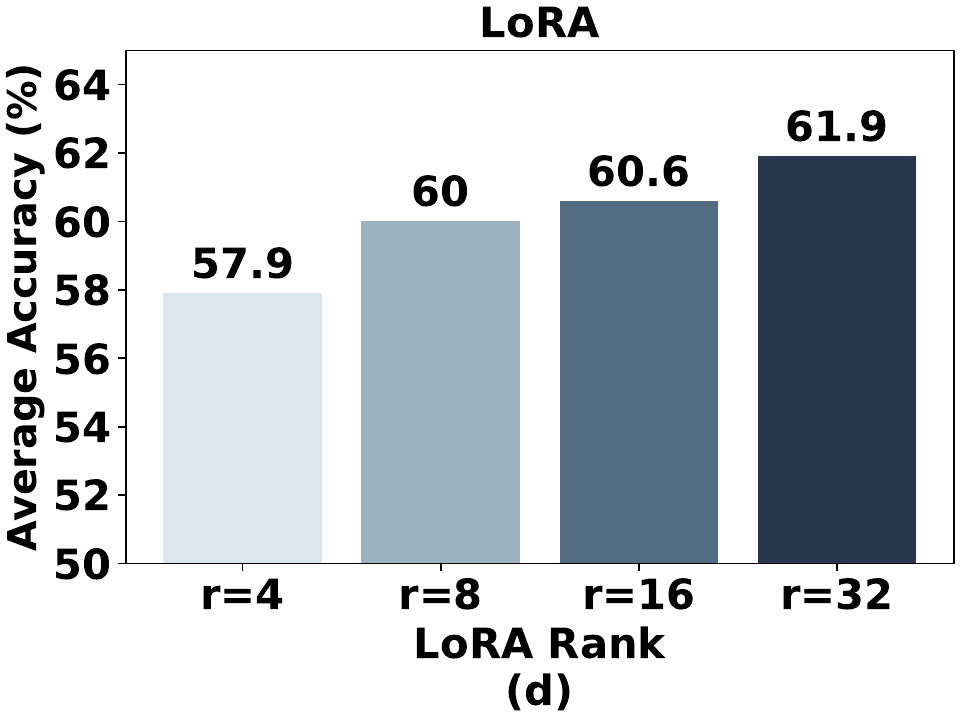} \\
    \end{tabular}
    \vspace{-5pt}
    \caption{The average accuracy of different variable settings on math reasoning datasets. Where "vt" refers to the number of virtual tokens, "bn" denotes the bottleneck size, while "r" is the LoRA rank. }
    \label{fig:configuration}
    \vspace{-10pt}
\end{figure}

\begin{table*}[t]\centering
\small
\begin{tabular}{llccccccc}
\hline
LLM & Method & MultiArith  &GSM8K &AddSub &AQuA &SingleEq &SVAMP & Avg \\\hline

GPT-3.5$_\text{175B}$ & - &  {83.8}  & \textbf{56.4}& {85.3}& \textbf{38.9} & {88.1} & \textbf{69.9} & \textbf{70.4}\\
\hline

\multirow{4}{*}{BLOOMz$_\text{7B}$}& Prefix & 68.8&	13.8&	47.1&	12.5&	49.4&	24.1&	36.0 \\
& Series  &80.7&	14.3&	72.6&	20.5&	69.3&	38.1&	49.3 \\
& Parallel & 85.8&	18.5&	77.7&	18.9&	74.8&	36.4&	52.0\\
& LoRA &82.8&	17.4&	72.4&	21.3&	69.9&	41.0&	50.8 \\

\hline

\multirow{4}{*}{GPT-J$_\text{6B}$}& Prefix &  74.5&	16.0&	65.6&	14.7&	61.4&	31.0&	43.9\\
& Series  & 91.7&	19.5&	85.8&	15.0&	81.7&	43.6&	56.2\\
& Parallel &92.2&	18.9&	83.8&	17.9&	80.7&	41.1&	55.8 \\
& LoRA &90.7&	23.0&	84.1&	16.1&	84.1&	46.0&	57.3 \\

\hline

\multirow{4}{*}{LLaMA$_\text{7B}$}& Prefix &63.2&	24.4&	57.0&	14.2&	55.3&	38.1&	42.0  \\
& Series  &92.8&	33.3&	80.0&	15.0&	83.5&	52.3&	59.5 \\
& Parallel &94.5&	35.3&	86.6&	18.1&	86.0&	49.6&	61.7 \\
& LoRA &\textbf{95.0}&	37.5&	83.3&	18.9&	84.4&	52.1&	61.9 \\
\hline

\multirow{4}{*}{LLaMA$_\text{13B}$}& Prefix &72.2&	31.1&	56.0&	15.7&	62.8&	41.4&	46.5  \\
& Series  &93.0&	44.0&	80.5&	22.0&	87.6&	50.8&	63.0 \\
& Parallel &94.3&	43.3&	83.0&	20.5&	89.6&	55.7&	64.4\\
& LoRA &94.8&	47.5&	\textbf{87.3}&	18.5&	\textbf{89.8}&	54.6&	65.4 \\

\hline
\end{tabular}
\vspace{-3pt}
\caption{Accuracy comparison of LLMs with different adapters on six math reasoning datasets. We use GPT-3.5 \texttt{text-Davinci-003} for Zero-shot CoT \cite{kojima2022large} as the baseline. 
}
\vspace{-15pt}
\label{tab:math_results}

\end{table*}

In order to determine the optimal configuration of various adapters, we conduct an analysis of the most crucial variable for each type of the PEFT methods. We compare the average accuracy on math reasoning datasets. The placement of adapters follows the optimal settings derived from the placement analysis. Regarding Prefix-tuning, we assess the performance with different numbers of virtual tokens ($vt$) set at $[10, 20, 30, 40]$. For Series and Parallel Adapters, we evaluate the impact of the bottleneck size ($bn$) with values of $[64, 128, 256, 512]$. For LoRA, we examine the influence of different rank values ($r$) at $[4, 8, 16, 32]$. The detailed results for each dataset can be found in Appendix \ref{configuration_analysis}. Figure \ref{fig:configuration} presents the average accuracy of different variables on math reasoning datasets. It can be noted that when the number of virtual tokens in Prefix-Tuning is set to 10, Prefix-Tuning attains an average accuracy of $42.0\%$ on math reasoning datasets. By configuring the bottleneck dimension to 256, Series and Parallel Adapter demonstrate the highest level of performance. However, when the bottleneck size is increased to 512, the accuracy of both Series and Parallel Adapter decreases. The typical setting for LoRA rank is set to 8, but we have discovered that a larger rank can enhance the performance of LoRA. When the rank is increased from 8 to 32, the average accuracy of LoRA increases from $60.0\%$ $61.9\%$. 

Based on our comprehensive placement and configuration analysis, we have determined the optimal settings for each adapter, which will be consistently employed throughout the subsequent experiments. 
\begin{itemize}
    \item \textbf{For Prefix-Tuning, we establish the number of virtual tokens at 10.
    \vspace{-5pt}
    \item For Series and Parallel Adapter, we seamlessly incorporate them into the MLP layers, configuring the bottleneck size to 256.
    \vspace{-15pt}
    \item Regarding LoRA, we seamlessly integrate it into both the Multi-head Attention layers and the MLP layers with rank 32.}
\end{itemize}


\subsection{Arithmetic Reasoning}

\begin{table*}[t]\centering
\small
\begin{tabular}{llccccccccc}
\hline
LLM & Method & BoolQ  &PIQA &SIQA &HellaSwag &WinoGrande &ARC-e &ARC-c &OBQA &Avg \\\hline

GPT-3$_\text{175B}$ &-& 60.5&	81.0&	-&	78.9&	70.2&	68.8&	51.4&	57.6& -\\
PaLM$_\text{540B}$ &-& \textbf{88.0}&	82.3&	-	&83.4&	81.1&	76.6&	53.0&	53.4& - \\
ChatGPT & - & 73.1&	\textbf{85.4}&	68.5&	78.5&	66.1&	\textbf{89.8}&	\textbf{79.9}&	74.8&	77.0 \\
\hline

\multirow{4}{*}{BLOOMz$_\text{7B}$}& Prefix &45.6&	53.7&	46.3&	26.7&	49.5&	52.1&	39.7&	44.3&	44.7  \\
& Series  &65.4&	70.4&	73.6&	53.4&	69.3&	72.3&	55.9&	68.0&	66.0 \\
& Parallel &64.1&	71.5&	72.1&	52.9&	67.0&	70.5&	54.7&	69.6&	65.3 \\
& LoRA &65.9&	75.3&	74.5&	57.3&	72.5&	74.6&	57.8&	73.4&	68.9 \\

\hline

\multirow{4}{*}{GPT-J$_\text{6B}$}& Prefix & 63.1&	66.9&	68.7&	34.4&	64.5&	64.4&	46.8&	59.0&	58.5 \\
& Series  &62.1&	63.5&	72.3&	30.6&	68.0&	63.9&	48.1&	63.8&	59.0 \\
& Parallel &62.2&	69.7&	70.0&	41.7&	65.0&	60.2&	44.6&	58.2&	59.0 \\
& LoRA &62.4&	68.6&	49.5&	43.1&	57.3&	43.4&	31.0&	46.6&	50.2 \\

\hline

\multirow{4}{*}{LLaMA$_\text{7B}$}& Prefix &64.3&	76.8&	73.9&	42.1&	72.1&	72.9&	54.0&	60.6&	64.6  \\
& Series  &63.0&	79.2&	76.3&	67.9&	75.7&	74.5&	57.1&	72.4&	70.8 \\
& Parallel &67.9&	76.4&	78.8&	69.8&	78.9&	73.7&	57.3&	75.2&	72.3 \\
& LoRA &68.9&	80.7&	77.4&	78.1&	78.8&	77.8&	61.3&	74.8&	74.7 \\
\hline

\multirow{4}{*}{LLaMA$_\text{13B}$}& Prefix &65.3&	75.4&	72.1&	55.2&	68.6&	79.5&	62.9&	68.0&	68.4  \\
& Series  &71.8&	83.0&	79.2&	88.1&	82.4&	82.5&	67.3&	81.8&	79.5 \\
& Parallel &72.5&	84.8&	79.8&	\textbf{92.1}&	\textbf{84.7}&	84.2&	71.2&	\textbf{82.4}&	\textbf{81.5} \\
& LoRA &72.1&	83.5&	\textbf{80.5}&	90.5&	83.7&	82.8&	68.3&	\textbf{82.4}&	80.5 \\

\hline
\end{tabular}
\vspace{-3pt}
\caption{Accuracy comparison of LLMs with different adapters on eight
commonsense reasoning datasets. The ChatGPT results are obtained by Zero-shot CoT with \texttt{gpt-3.5-turbo} API. 
}
\vspace{-10pt}
\label{tab:commonsense_results}

\end{table*}

In order to evaluate the effectiveness of adapters on the Arithmetic Reasoning task, we conducted a study where adapters are fine-tuned on the Math10K dataset and subsequently evaluated on six different math reasoning datasets. As our baseline, we utilize the GPT-3.5 model, specifically the \texttt{text-Davinci-003} variant, for Zero-shot CoT according to \citet{kojima2022large}. The results of the GPT-3.5 model can be found in \citet{wang2023plan}. Table \ref{tab:math_results} reports the performance of different PEFT methods and the baseline. On average, the GPT-3.5 model (175B) outperforms adapter-based PEFT LLMs in terms of accuracy. However, for simpler math reasoning datasets such as MultiArith, AddSub, and SingleEq, adapter-based methods like LLaMA-13B with LoRA outperform GPT-3.5. Notably, LLaMA-13B with LoRA achieves an average accuracy of 65.4\%, which is approximately 92.8\% of the performance exhibited by GPT-3.5. This suggests that with sufficient task-specific training data, adapter-based PEFT of smaller LLMs has the potential to achieve performance comparable to that of extremely large language models. The utilization of adapter-based PEFT yields superior performance by smaller language models compared to GPT-3.5 specifically in simpler tasks such as MultiArith, AddSub, and SingleEq. However, challenges persist in more complex tasks like GSM8K and SVAMP, which require a higher level of language comprehension and proficiency from the underlying base model, thereby resulting in a discernible performance gap.
Regarding the different adapters employed, LoRA achieves remarkable performance while utilizing significantly fewer trainable parameters. This implies that excessive learnable parameters may not be necessary for task-specific fine-tuning. Overall, these findings demonstrate the potential for adapter-based PEFT of smaller LLMs to achieve high performance on specific tasks with few trainable parameters. 

\begin{table*}[t]\centering
\small
\begin{tabular}{lp{13cm}}\hline
Question &  Mr Boarden is remodeling his bathroom. For every square foot, he needs 24 mosaic tiles. How many mosaic tiles would Mr Boarden need to cover two thirds of his 36 sq ft bathroom? \\
\hline
ChatGPT & 
1. Find two-thirds of 36: Multiply 36 by 2/3. \textbf{(36 * 2/3) = 24 square feet}

2.Determine the number of mosaic tiles needed per square foot: Given that Mr. Boarden needs 24 mosaic tiles per square foot, multiply the result from step 1 by 24. \textbf{(24 * 24) = 576 mosaic tiles}

Therefore, Mr. Boarden would need 576 mosaic tiles to cover two-thirds of his 36 square foot bathroom.
\\ \hline

LLaMA$_\text{13B}$-Prefix & 1 square foot = 4 square inches, 1 square inch = 16 square centimeters. So, 1 sq ft = 16 sq cm. Therefore, 36 sq ft = 36 x 16 sq cm = 576 sq cm.
To cover two-thirds of the bathroom, Mr Boarden needs to cover 576 sq cm x 2/3 = 384 sq cm.
Since 1 sq cm = 1 mosaic tile, Mr Boarden needs 384 mosaic tiles.
Therefore, Mr Boarden needs 384 mosaic tiles to cover two-thirds of his 36 sq ft bathroom. \\ \hline

LLaMA$_\text{13B}$-Series & 1. Find the area of the bathroom: 36 sq ft

2. Find two-thirds of the area: \textbf{2/3 x 36 sq ft = 24 sq ft}

3. Multiply the area by the number of tiles needed per square foot: \textbf{24 sq ft x 24 tiles/sq ft = 576 tiles}

4. Therefore, Mr Boarden would need 576 mosaic tiles to cover two-thirds of his 36 sq ft bathroom. \\ \hline

LLaMA$_\text{13B}$-Parallel &  1. Calculate the total area of the bathroom: 36 sq ft

2. Determine how much area Mr Boarden wants to cover with mosaic tiles: \textbf{2/3 x 36 sq ft = 24 sq ft}

3. Calculate the number of mosaic tiles needed: \textbf{24 sq ft x 24} mosaic tiles per sq ft = 600 mosaic tiles

Therefore, Mr Boarden would need 600 mosaic tiles to cover two thirds of his 36 sq ft bathroom.
\\\hline
LLaMA$_\text{13B}$-LoRA &  1. Calculate the total area of the bathroom: 36 sq ft

2. Determine how much area needs to be covered: \textbf{2/3 x 36 sq ft = 24 sq ft}

3. Calculate the number of mosaic tiles needed: \textbf{24 sq ft x 24 mosaic tiles per sq ft = 576 mosaic tiles}

Therefore, Mr Boarden would need 576 mosaic tiles to cover two thirds of his 36 sq ft bathroom.  \\\hline

\hline
\end{tabular}
\vspace{-3pt}
\caption{An example randomly sampled from GSM8K. The outputs of ChatGPT and LLaMA-13B with different PEFT methods.}
\label{tab:case_study}
\vspace{-10pt}
\end{table*}


\subsection{Commonsense Reasoning}

Additionally, we assess the efficacy of various PEFT methods for commonsense reasoning tasks. The adapters undergo fine-tuning using the Commonsense170K dataset. Our baseline models for commonsense reasoning include GPT-3 (175B), PaLM (540B), and ChatGPT. The results for GPT-3 and PaLM can be found in the study by \citet{llama}. To evaluate ChatGPT's performance in commonsense reasoning, we employ the \texttt{gpt-3.5-turbo} API with a zero-shot CoT. The zero-shot CoT prompts align with the template used for collecting our commonsense fine-tuning dataset, as outlined in Appendix \ref{sec:commonsense_templates}. Table \ref{tab:commonsense_results} presents the performance of the PEFT methods utilizing different LLMs  alongside the baselines. Remarkably, LLaMA-13B with Series Adapter, Parallel Adapter, and LoRA outperform all the baselines, including ChatGPT, which has been hailed as the most impressive LLM to date. LLaMA-13B with Parallel Adapter achieves an average accuracy of 81.5\%, representing a 4.5\% improvement over ChatGPT. It is worth noting that all the training sets from the commonsense reasoning datasets are included in the fine-tuning data Commonsense170K. Furthermore, we observe that the performance of the PEFT methods is influenced by the underlying capabilities of the base models. LLaMA-7B and LLaMA-13B demonstrate superior commonsense reasoning abilities compared to the BLOOMz and GPT-J models.

\subsection{ID and OOD Analysis}
When comparing the performance of PEFT methods on math reasoning and commonsense reasoning tasks, we can observe that PEFT methods exhibit more remarkable results in the realm of commonsense reasoning. Moving forward, we will analyze the factors contributing to this phenomenon from both the in-distribution (ID) and out-of-distribution (OOD) perspectives. In the context of commonsense reasoning, the fine-tuning data set, Commonsense170K, encompasses all the training sets from the commonsense reasoning datasets. Notably, PEFT methods have demonstrated the ability to outperform ChatGPT. This observation implies that, by utilizing ID fine-tuning data, smaller language models like LLaMA-13B could surpass larger language models such as ChatGPT and PaLM in specific downstream tasks. However, when considering math reasoning tasks, the fine-tuning data set, Math10K, only includes the training sets of GSM8K and AQuA. In this regard, it has been observed that PEFT methods, particularly LLaMA-13B with LoRA, exhibit superior performance compared to GPT-3.5 on MultiArith, AddSub, and SingleEq. These findings suggest that PEFT methods can enhance the math reasoning abilities of LLMs and can be successfully applied to OOD datasets. Nonetheless, when evaluating the performance of PEFT methods on the ID datasets GSM8K and AQuA, a performance gap is still evident compared to GPT-3.5. This discrepancy is likely due to the higher complexity of GSM8K and AQuA datasets in terms of math reasoning, while the reasoning capabilities of smaller LLMs remain limited. Consequently, identifying strategies to improve the performance of PEFT methods on complex math reasoning tasks represents a potential avenue for future research.

\section{Qualitative Study}

The previous sections have presented the quantitative analysis. In this section, we will provide qualitative examples to demonstrate the quality of outputs from different models. Table \ref{tab:case_study} displays a randomly selected question from GSM8K along with the outputs of ChatGPT and LLaMA-13B models using various PEFT methods. More detailed examples can be found in Appendix \ref{sec:examples}. ChatGPT demonstrates a comprehensive understanding of the question and generates two steps, "(36 * 2/3) = 24 square feet" and "(24 * 24) = 576 mosaic tiles," effectively solving the problem. However, the language understanding ability of LLaMA-13B-Prefix models is limited, leading LLaMA-13B-Prefix to take the wrong direction in the first step. On the other hand, LLaMA-13B with Series Adapter produces a high-quality answer by providing the crucial two steps and performing the correct calculations to obtain the accurate result. Interestingly, LLaMA-13B-Parallel and LLaMA-13B-LoRA generate almost identical rationales. However, LLaMA-13B-Parallel produces an incorrect answer due to a calculation error, stating "24 sq ft x 24 mosaic tiles per sq ft = 600 mosaic tiles". In general, when equipped with task-specific fine-tuning data, smaller language models like LLaMA-13B can generate impressive, high-quality answers that are comparable to those produced by ChatGPT.

\section{Conclusion}
 In this paper, we develop a user-friendly framework, LLM-Adapter, seamlessly integrates diverse adapters into LLMs, empowering researchers to implement adapter-based PEFT methods for a wide range of tasks. To evaluate different PEFT methods on downstream tasks, we construct two high-quality fine-tuning datasets to enhance PEFT performance on math reasoning and commonsense reasoning tasks. By utilizing the LLM-Adapter toolkit and the constructed fine-tuning datasets, we conduct a comprehensive empirical study and find the answer of research questions on the optimal placement and configuration of different PEFT methods, the impact of adapter architectures, and the influence of ID and OOD scenarios. We hope this work will encourage further research on PEFT methods for LLMs. 

\section{Limitations}
There are two limitations to this work. Firstly, due to constrained computing resources, we were unable to evaluate the performance of larger language models such as LLaMA-33B and LLaMA-65B. It is anticipated that these larger models, possessing enhanced language understanding capabilities, would yield superior performance. Secondly, this paper does not delve into the exploration of combining different adapters. Given the extensive search space associated with the combination of various PEFT methods, we intend to explore this direction in future research endeavors.


\bibliography{anthology,custom}

\begin{thebibliography}{44}
\expandafter\ifx\csname natexlab\endcsname\relax\def\natexlab#1{#1}\fi

\bibitem[{Aghajanyan et~al.(2020)Aghajanyan, Zettlemoyer, and
  Gupta}]{intrinsic_said}
Armen Aghajanyan, Luke Zettlemoyer, and Sonal Gupta. 2020.
\newblock Intrinsic dimensionality explains the effectiveness of language model
  fine-tuning.
\newblock In \emph{Annual Meeting of the Association for Computational
  Linguistics}.

\bibitem[{Bisk et~al.(2020)Bisk, Zellers, Bras, Gao, and Choi}]{piqa}
Yonatan Bisk, Rowan Zellers, Ronan~Le Bras, Jianfeng Gao, and Yejin Choi. 2020.
\newblock Piqa: Reasoning about physical commonsense in natural language.
\newblock In \emph{Thirty-Fourth AAAI Conference on Artificial Intelligence}.

\bibitem[{Chen et~al.(2023)Chen, Zhang, Shi, Li, Smola, and Yang}]{s4_model}
Jiaao Chen, Aston Zhang, Xingjian Shi, Mu~Li, Alex Smola, and Diyi Yang. 2023.
\newblock Parameter-efficient fine-tuning design spaces.
\newblock \emph{arXiv preprint arXiv:2301.01821}.

\bibitem[{Clark et~al.(2019)Clark, Lee, Chang, Kwiatkowski, Collins, and
  Toutanova}]{boolq}
Christopher Clark, Kenton Lee, Ming-Wei Chang, Tom Kwiatkowski, Michael
  Collins, and Kristina Toutanova. 2019.
\newblock \href {https://doi.org/10.18653/v1/N19-1300} {{B}ool{Q}: Exploring
  the surprising difficulty of natural yes/no questions}.
\newblock In \emph{Proceedings of the 2019 Conference of the North {A}merican
  Chapter of the Association for Computational Linguistics: Human Language
  Technologies, Volume 1 (Long and Short Papers)}, pages 2924--2936,
  Minneapolis, Minnesota. Association for Computational Linguistics.

\bibitem[{Clark et~al.(2018)Clark, Cowhey, Etzioni, Khot, Sabharwal, Schoenick,
  and Tafjord}]{arc}
Peter Clark, Isaac Cowhey, Oren Etzioni, Tushar Khot, Ashish Sabharwal, Carissa
  Schoenick, and Oyvind Tafjord. 2018.
\newblock Think you have solved question answering? try arc, the ai2 reasoning
  challenge.
\newblock \emph{arXiv:1803.05457v1}.

\bibitem[{Cobbe et~al.(2021)Cobbe, Kosaraju, Bavarian, Hilton, Nakano, Hesse,
  and Schulman}]{gsm8k}
Karl Cobbe, Vineet Kosaraju, Mohammad Bavarian, Jacob Hilton, Reiichiro Nakano,
  Christopher Hesse, and John Schulman. 2021.
\newblock Training verifiers to solve math word problems.
\newblock \emph{arXiv preprint arXiv:2110.14168}.

\bibitem[{Devlin et~al.(2018)Devlin, Chang, Lee, and
  Toutanova}]{devlin2018bert}
Jacob Devlin, Ming-Wei Chang, Kenton Lee, and Kristina Toutanova. 2018.
\newblock Bert: Pre-training of deep bidirectional transformers for language
  understanding.
\newblock \emph{arXiv preprint arXiv:1810.04805}.

\bibitem[{Edalati et~al.(2022)Edalati, Tahaei, Kobyzev, Nia, Clark, and
  Rezagholizadeh}]{krona}
Ali Edalati, Marzieh~S. Tahaei, Ivan Kobyzev, V.~Nia, James~J. Clark, and Mehdi
  Rezagholizadeh. 2022.
\newblock Krona: Parameter efficient tuning with kronecker adapter.
\newblock \emph{ArXiv}, abs/2212.10650.

\bibitem[{Fu et~al.(2021)Fu, Huang, Chen, Tian, and Zhao}]{lets}
Cheng Fu, Hanxian Huang, Xinyun Chen, Yuandong Tian, and Jishen Zhao. 2021.
\newblock \href {https://proceedings.mlr.press/v139/fu21a.html}
  {Learn-to-share: A hardware-friendly transfer learning framework exploiting
  computation and parameter sharing}.
\newblock In \emph{Proceedings of the 38th International Conference on Machine
  Learning}, volume 139 of \emph{Proceedings of Machine Learning Research},
  pages 3469--3479. PMLR.

\bibitem[{He et~al.(2021)He, Zhou, Ma, Berg-Kirkpatrick, and Neubig}]{mam}
Junxian He, Chunting Zhou, Xuezhe Ma, Taylor Berg-Kirkpatrick, and Graham
  Neubig. 2021.
\newblock Towards a unified view of parameter-efficient transfer learning.
\newblock \emph{arXiv preprint arXiv:2110.04366}.

\bibitem[{He et~al.(2022{\natexlab{a}})He, Zhou, Ma, Berg-Kirkpatrick, and
  Neubig}]{parallel_adapter}
Junxian He, Chunting Zhou, Xuezhe Ma, Taylor Berg-Kirkpatrick, and Graham
  Neubig. 2022{\natexlab{a}}.
\newblock \href {https://openreview.net/forum?id=0RDcd5Axok} {Towards a unified
  view of parameter-efficient transfer learning}.
\newblock In \emph{International Conference on Learning Representations}.

\bibitem[{He et~al.(2022{\natexlab{b}})He, Ding, Dong, Zhang, and
  Tao}]{sparseadapter}
Shwai He, Liang Ding, Daize Dong, Jeremy Zhang, and Dacheng Tao.
  2022{\natexlab{b}}.
\newblock \href {https://aclanthology.org/2022.findings-emnlp.160}
  {{S}parse{A}dapter: An easy approach for improving the parameter-efficiency
  of adapters}.
\newblock In \emph{Findings of the Association for Computational Linguistics:
  EMNLP 2022}, pages 2184--2190, Abu Dhabi, United Arab Emirates. Association
  for Computational Linguistics.

\bibitem[{Henderson et~al.(2021)Henderson, Ruder et~al.}]{compacter}
James Henderson, Sebastian Ruder, et~al. 2021.
\newblock Compacter: Efficient low-rank hypercomplex adapter layers.
\newblock In \emph{Advances in Neural Information Processing Systems}.

\bibitem[{Hosseini et~al.(2014)Hosseini, Hajishirzi, Etzioni, and
  Kushman}]{addsub}
Mohammad~Javad Hosseini, Hannaneh Hajishirzi, Oren Etzioni, and Nate Kushman.
  2014.
\newblock Learning to solve arithmetic word problems with verb categorization.
\newblock In \emph{EMNLP}, pages 523--533.

\bibitem[{Houlsby et~al.(2019)Houlsby, Giurgiu, Jastrzebski, Morrone,
  de~Laroussilhe, Gesmundo, Attariyan, and Gelly}]{adapters}
Neil Houlsby, Andrei Giurgiu, Stanislaw Jastrzebski, Bruna Morrone, Quentin
  de~Laroussilhe, Andrea Gesmundo, Mona Attariyan, and Sylvain Gelly. 2019.
\newblock Parameter-efficient transfer learning for nlp.
\newblock In \emph{International Conference on Machine Learning}.

\bibitem[{Hu et~al.(2021)Hu, Shen, Wallis, Allen-Zhu, Li, Wang, and
  Chen}]{lora}
Edward~J. Hu, Yelong Shen, Phillip Wallis, Zeyuan Allen-Zhu, Yuanzhi Li, Shean
  Wang, and Weizhu Chen. 2021.
\newblock Lora: Low-rank adaptation of large language models.
\newblock \emph{ArXiv}, abs/2106.09685.

\bibitem[{Kojima et~al.(2022)Kojima, Gu, Reid, Matsuo, and
  Iwasawa}]{kojima2022large}
Takeshi Kojima, Shixiang~Shane Gu, Machel Reid, Yutaka Matsuo, and Yusuke
  Iwasawa. 2022.
\newblock Large language models are zero-shot reasoners.
\newblock \emph{arXiv preprint arXiv:2205.11916}.

\bibitem[{Koncel-Kedziorski et~al.(2015)Koncel-Kedziorski, Hajishirzi,
  Sabharwal, Etzioni, and Ang}]{singleeq}
Rik Koncel-Kedziorski, Hannaneh Hajishirzi, Ashish Sabharwal, Oren Etzioni, and
  Siena~Dumas Ang. 2015.
\newblock Parsing algebraic word problems into equations.
\newblock \emph{Transactions of the Association for Computational Linguistics},
  3:585--597.

\bibitem[{Koncel-Kedziorski et~al.(2016)Koncel-Kedziorski, Roy, Amini, Kushman,
  and Hajishirzi}]{mawps}
Rik Koncel-Kedziorski, Subhro Roy, Aida Amini, Nate Kushman, and Hannaneh
  Hajishirzi. 2016.
\newblock \href {https://aclanthology.org/N16-1136} {{MAWPS}: A math word
  problem repository}.
\newblock In \emph{Proceedings of NAACL}, pages 1152--1157.

\bibitem[{Lester et~al.(2021)Lester, Al-Rfou, and Constant}]{prompt_tuning}
Brian Lester, Rami Al-Rfou, and Noah Constant. 2021.
\newblock The power of scale for parameter-efficient prompt tuning.
\newblock \emph{ArXiv}, abs/2104.08691.

\bibitem[{Li and Liang(2021)}]{prefix}
Xiang~Lisa Li and Percy Liang. 2021.
\newblock \href {https://doi.org/10.18653/v1/2021.acl-long.353} {Prefix-tuning:
  Optimizing continuous prompts for generation}.
\newblock In \emph{Proceedings of the 59th Annual Meeting of the Association
  for Computational Linguistics and the 11th International Joint Conference on
  Natural Language Processing (Volume 1: Long Papers)}, pages 4582--4597,
  Online. Association for Computational Linguistics.

\bibitem[{Ling et~al.(2017)Ling, Yogatama, Dyer, and Blunsom}]{aqua}
Wang Ling, Dani Yogatama, Chris Dyer, and Phil Blunsom. 2017.
\newblock Program induction by rationale generation: Learning to solve and
  explain algebraic word problems.
\newblock In \emph{Proceedings of the 55th Annual Meeting of the Association
  for Computational Linguistics (Volume 1: Long Papers)}, pages 158--167.

\bibitem[{Mangrulkar et~al.(2022)Mangrulkar, Gugger, Debut, Belkada, and
  Paul}]{peft}
Sourab Mangrulkar, Sylvain Gugger, Lysandre Debut, Younes Belkada, and Sayak
  Paul. 2022.
\newblock Peft: State-of-the-art parameter-efficient fine-tuning methods.
\newblock \url{https://github.com/huggingface/peft}.

\bibitem[{Mao et~al.(2021)Mao, Mathias, Hou, Almahairi, Ma, Han, tau Yih, and
  Khabsa}]{unipelt}
Yuning Mao, Lambert Mathias, Rui Hou, Amjad Almahairi, Hao Ma, Jiawei Han, Wen
  tau Yih, and Madian Khabsa. 2021.
\newblock Unipelt: A unified framework for parameter-efficient language model
  tuning.
\newblock \emph{ArXiv}, abs/2110.07577.

\bibitem[{Muennighoff et~al.(2022)Muennighoff, Wang, Sutawika, Roberts,
  Biderman, Scao, Bari, Shen, Yong, Schoelkopf et~al.}]{bloom}
Niklas Muennighoff, Thomas Wang, Lintang Sutawika, Adam Roberts, Stella
  Biderman, Teven~Le Scao, M~Saiful Bari, Sheng Shen, Zheng-Xin Yong, Hailey
  Schoelkopf, et~al. 2022.
\newblock Crosslingual generalization through multitask finetuning.
\newblock \emph{arXiv preprint arXiv:2211.01786}.

\bibitem[{OpenAI(2022)}]{openai-chatgpt-2022}
OpenAI. 2022.
\newblock Introducing chatgpt.
\newblock \url{https://openai.com/blog/chatgpt}.

\bibitem[{OpenAI(2023)}]{openai-gpt4-2023}
OpenAI. 2023.
\newblock {GPT-4} technical report.
\newblock \emph{CoRR}, abs/2303.08774.

\bibitem[{Patel et~al.(2021)Patel, Bhattamishra, and Goyal}]{svamp}
Arkil Patel, Satwik Bhattamishra, and Navin Goyal. 2021.
\newblock \href {https://aclanthology.org/2021.naacl-main.168} {Are {NLP}
  models really able to solve simple math word problems?}
\newblock In \emph{Proceedings of NAACL}, pages 2080--2094.

\bibitem[{Pfeiffer et~al.(2020)Pfeiffer, Vulic, Gurevych, and Ruder}]{madx}
Jonas Pfeiffer, Ivan Vulic, Iryna Gurevych, and Sebastian Ruder. 2020.
\newblock Mad-x: An adapter-based framework for multi-task cross-lingual
  transfer.
\newblock In \emph{Conference on Empirical Methods in Natural Language
  Processing}.

\bibitem[{Qin et~al.(2023)Qin, Zhang, Zhang, Chen, Yasunaga, and
  Yang}]{qin-chatgpt-2023}
Chengwei Qin, Aston Zhang, Zhuosheng Zhang, Jiaao Chen, Michihiro Yasunaga, and
  Diyi Yang. 2023.
\newblock Is chatgpt a general-purpose natural language processing task solver?
\newblock \emph{arXiv preprint arXiv:2302.06476}.

\bibitem[{Qin et~al.(2021)Qin, Wang, Su, Lin, Ding, Yi, Chen, Liu, Li, Hou
  et~al.}]{ipt}
Yujia Qin, Xiaozhi Wang, Yusheng Su, Yankai Lin, Ning Ding, Jing Yi, Weize
  Chen, Zhiyuan Liu, Juanzi Li, Lei Hou, et~al. 2021.
\newblock Exploring universal intrinsic task subspace via prompt tuning.
\newblock \emph{arXiv e-prints}, pages arXiv--2110.

\bibitem[{Roy and Roth(2016)}]{mutli_arith}
Subhro Roy and Dan Roth. 2016.
\newblock Solving general arithmetic word problems.
\newblock \emph{arXiv preprint arXiv:1608.01413}.

\bibitem[{Sakaguchi et~al.(2021)Sakaguchi, Bras, Bhagavatula, and
  Choi}]{winogrande}
Keisuke Sakaguchi, Ronan~Le Bras, Chandra Bhagavatula, and Yejin Choi. 2021.
\newblock Winogrande: An adversarial winograd schema challenge at scale.
\newblock \emph{Communications of the ACM}, 64(9):99--106.

\bibitem[{Sap et~al.(2019)Sap, Rashkin, Chen, LeBras, and Choi}]{siqa}
Maarten Sap, Hannah Rashkin, Derek Chen, Ronan LeBras, and Yejin Choi. 2019.
\newblock Socialiqa: Commonsense reasoning about social interactions.
\newblock \emph{arXiv preprint arXiv:1904.09728}.

\bibitem[{Shen et~al.(2023)Shen, Song, Tan, Li, Lu, and Zhuang}]{hugginggpt}
Yongliang Shen, Kaitao Song, Xu~Tan, Dongsheng Li, Weiming Lu, and Yueting
  Zhuang. 2023.
\newblock Hugginggpt: Solving {AI} tasks with chatgpt and its friends in
  huggingface.
\newblock \emph{CoRR}, abs/2303.17580.

\bibitem[{Sung et~al.(2022)Sung, Cho, and Bansal}]{ladder_side_tuning}
Yi-Lin Sung, Jaemin Cho, and Mohit Bansal. 2022.
\newblock Lst: Ladder side-tuning for parameter and memory efficient transfer
  learning.
\newblock \emph{ArXiv}, abs/2206.06522.

\bibitem[{Taori et~al.(2023)Taori, Gulrajani, Zhang, Dubois, Li, Guestrin,
  Liang, and Hashimoto}]{alpaca}
Rohan Taori, Ishaan Gulrajani, Tianyi Zhang, Yann Dubois, Xuechen Li, Carlos
  Guestrin, Percy Liang, and Tatsunori~B. Hashimoto. 2023.
\newblock Stanford alpaca: An instruction-following llama model.
\newblock \url{https://github.com/tatsu-lab/stanford_alpaca}.

\bibitem[{Touvron et~al.(2023)Touvron, Lavril, Izacard, Martinet, Lachaux,
  Lacroix, Rozi{\`e}re, Goyal, Hambro, Azhar et~al.}]{llama}
Hugo Touvron, Thibaut Lavril, Gautier Izacard, Xavier Martinet, Marie-Anne
  Lachaux, Timoth{\'e}e Lacroix, Baptiste Rozi{\`e}re, Naman Goyal, Eric
  Hambro, Faisal Azhar, et~al. 2023.
\newblock Llama: Open and efficient foundation language models.
\newblock \emph{arXiv preprint arXiv:2302.13971}.

\bibitem[{Vaswani et~al.(2017)Vaswani, Shazeer, Parmar, Uszkoreit, Jones,
  Gomez, Kaiser, and Polosukhin}]{vaswani2017attention}
Ashish Vaswani, Noam Shazeer, Niki Parmar, Jakob Uszkoreit, Llion Jones,
  Aidan~N Gomez, {\L}ukasz Kaiser, and Illia Polosukhin. 2017.
\newblock Attention is all you need.
\newblock In \emph{Advances in neural information processing systems}, pages
  5998--6008.

\bibitem[{Vu et~al.(2021)Vu, Lester, Constant, Al-Rfou, and Cer}]{spot}
Tu~Vu, Brian Lester, Noah Constant, Rami Al-Rfou, and Daniel Cer. 2021.
\newblock Spot: Better frozen model adaptation through soft prompt transfer.
\newblock \emph{arXiv preprint arXiv:2110.07904}.

\bibitem[{Wang and Komatsuzaki(2021)}]{gpt-j}
Ben Wang and Aran Komatsuzaki. 2021.
\newblock {GPT-J-6B: A 6 Billion Parameter Autoregressive Language Model}.
\newblock \url{https://github.com/kingoflolz/mesh-transformer-jax}.

\bibitem[{Wang et~al.(2023)Wang, Xu, Lan, Hu, Lan, Lee, and Lim}]{wang2023plan}
Lei Wang, Wanyu Xu, Yihuai Lan, Zhiqiang Hu, Yunshi Lan, Roy Ka-Wei Lee, and
  Ee-Peng Lim. 2023.
\newblock Plan-and-solve prompting: Improving zero-shot chain-of-thought
  reasoning by large language models.
\newblock \emph{arXiv preprint arXiv:2305.04091}.

\bibitem[{Wang et~al.(2022)Wang, Mukherjee, Liu, Gao, Awadallah, and
  Gao}]{adamix}
Yaqing Wang, Subhabrata Mukherjee, Xiaodong Liu, Jing Gao, Ahmed~Hassan
  Awadallah, and Jianfeng Gao. 2022.
\newblock Adamix: Mixture-of-adapter for parameter-efficient tuning of large
  language models.
\newblock \emph{ArXiv}, abs/2205.12410.

\bibitem[{Yunxiang et~al.(2023)Yunxiang, Zihan, Kai, Ruilong, and
  You}]{yunxiang2023chatdoctor}
Li~Yunxiang, Li~Zihan, Zhang Kai, Dan Ruilong, and Zhang You. 2023.
\newblock Chatdoctor: A medical chat model fine-tuned on llama model using
  medical domain knowledge.
\newblock \emph{arXiv preprint arXiv:2303.14070}.

\end{thebibliography}
\bibliographystyle{acl_natbib}

\clearpage

\appendix

\section{Appendix}

\subsection{Math Reasoning Prompt Templates}
\label{sec:math_templates}
We utilize ChatGPT to collect the math reasoning data for fine-tuning. Table \ref{tab:p_math} show the prompt template used to query ChatGPT. The expression "Please give the steps" is employed to guide ChatGPT to generate reasoning steps, thus, we can use the rationale information to fine-tune adapters. "Give the arabic numerals as the answer." is utilized to guide ChatGPT to generate arabic numbers as the final answer making it easier to extract the answer from the outputs.

\begin{table*}[t]\centering
\small
\begin{tabular}{lp{13cm}}\hline
Template &  [QUESTION] Please give the steps and the arabic numerals as the answer. \\
\hline
Prompt & Albert is wondering how much pizza he can eat in one day. He buys 2 large pizzas and 2 small pizzas. A large pizza has 16 slices and a small pizza has 8 slices. If he eats it all, how many pieces does he eat that day? 

Please give the steps and the arabic numerals as the answer. \\ \hline

ChatGPT Output & To solve the problem, we need to first find out how many slices are in 2 large pizzas and 2 small pizzas: 2 large pizzas x 16 slices per pizza = 32 slices

2 small pizzas x 8 slices per pizza = 16 slices. 

Next, we add the number of slices in each size pizza to find the total number of slices: 32 slices + 16 slices = 48 slices. 

Therefore, if Albert eats all 2 large pizzas and 2 small pizzas, he will eat 48 slices of pizza that day. 

Answer: 48 slices. \\

\hline
\end{tabular}
\caption{The prompt template used to collect math reasoning dataset for fine-tuning. An example from GSM8K is also included.}
\label{tab:p_math}
\end{table*}

\subsection{Commonsense Data Templates}
\label{sec:commonsense_templates}

As each dataset in the commonsense reasoning domain entails distinct tasks, we adopt a structured template by initially describing the task’s goal, followed by the corresponding content and answer. Table \ref{tab:p_commonsense} shows the templates used to collect commonsense reasoning data for fine-tuning. 

\begin{table*}[t]\centering
\small
\begin{tabular}{lp{13cm}}\hline
Dataset & Fine-tuning Data Template \\
\hline
BoolQ & Please answer the following question with true or false, question: [QUESTION]

Answer format: true/false

the correct answer is [ANSWER] \\ \hline

PIQA & Please choose the correct solution to the question: [QUESTION]

Solution1: [SOLUTION\_1]

Solution2: [SOLUTION\_2]

Answer format: solution1/solution2 

the correct answer is [ANSWER]\\ \hline
SIQA & Please choose the correct answer to the question: [QUESTION]

Answer1: [ANSWER\_1]

Answer2: [ANSWER\_2]

Answer3: [ANSWER\_3]

Answer format: answer1/answer2/answer3 

the correct answer is [ANSWER]\\ \hline
HellaSwag & Please choose the correct ending to complete the given sentence: [ACTIVITY\_lABEL]: [CONTEXT]

Ending1: [ENDING\_1] 

Ending2: [ENDING\_2] 

Ending3: [ENDING\_3] 

Ending4: [ENDING\_4] 

Answer format: ending1/ending2/ending3/ending4 

the correct answer is [ANSWER]\\ \hline
WinoGrande &  Please choose the correct answer to fill in the blank to complete the given sentence: [SENTENCE]

Option1: [OPTION\_1]

Option2: [OPTION\_2]

the correct answer is [ANSWER]\\ \hline
ARC-e\&ARC-c & Please choose the correct answer to the question: [QUESTION] 

Answer1: [ANSWER\_1]

Answer2: [ANSWER\_2]

Answer3: [ANSWER\_3]

Answer4: [ANSWER\_4]

Answer format: answer1/answer2/answer3/answer4

the correct answer is [ANSWER]
\\    \hline
OBQA &     Please choose the correct answer to the question: [QUESTION] 

Answer1: [ANSWER\_1]

Answer2: [ANSWER\_2]

Answer3: [ANSWER\_3]

Answer4: [ANSWER\_4]

Answer format: answer1/answer2/answer3/answer4

the correct answer is [ANSWER]  \\    \hline

\hline
\end{tabular}
\caption{The data template of each dataset used to create commonsense reasoning data for fine-tuning.}
\label{tab:p_commonsense}
\end{table*}

\subsection{Placement Analysis}
\label{placement_analysis}

Table \ref{tab:placement_results} shows the performance regarding the placement of adapters in various locations on math reasoning datasets. The fine-tuning dataset utilized for this study is Math10K. Meanwhile, the base models employed is LLaMA-7B. We can observe that for the Series Adapter, the best position is to place it after the MLP layers, achieving an average accuracy of $59.5\%$ on the math reasoning datasets. As for the Parallel Adapter, when we place it within the MLP layers, it achieves the best performance of $61.7\%$. Regarding LoRA, we need to insert it simultaneously into both the Multi-head Attention layers and MLP layers to achieve the best performance of $60\%$.

\begin{table*}[t]\centering
\begin{tabular}{llccccccc}
\hline
Model & Location & MultiArith  &GSM8K &AddSub &AQuA &SingleEq &SVAMP & Average \\\hline


\multirow{3}{*}{Series}& Attn & 92.3 & 32.0 & 80.0 & 16.9 & 80.5 & 47.9 & 58.3 \\
& MLP  & 92.8 & 33.3 & 80.0 & 15.0 & 83.5 & 52.3 & 59.5 \\
& Both & 94 & 29.8 & 84.1 & 17.3 & 83.5 & 45.1 & 59.0\\
\hline

\multirow{3}{*}{Parallel}& Attn & 94.5 & 33.5 & 83.0 & 17.3 & 80.5 & 46.9 & 59.3 \\
& MLP  &94.5& 35.3& \textbf{86.6}& \textbf{18.1}& \textbf{86.0}& \textbf{49.6}& \textbf{61.7 }\\
& Both &94.3& 30.2& 84.8& 17.7& 84.3& 47.2& 59.8\\

\hline

\multirow{3}{*}{LoRA}& Attn &94.2& 35.3& 79.7& 16.9& 78.7& 45.9& 58.5\\
& MLP & 95.8& 35.0& 80.0& 15.7& 81.7& 47.0& 59.2 \\
& Both & \textbf{96.2} & \textbf{35.6} & 80.5 & 15.7 & 82.3 & \textbf{49.6} & 60.0\\

\hline
\end{tabular}
\caption{An evaluation of the accuracy regarding the placement of adapters in various locations is conducted on math reasoning datasets. The fine-tuning dataset used for this analysis is Math10K. In this context, "Attn" refers to the multi-head attention layer, while "MLP" denotes the MLP layer. The base model employed for this study is LLaMA-7B.
}
\label{tab:placement_results}

\end{table*}

\subsection{Configuration Analysis}
\label{configuration_analysis}

Table \ref{tab:configuration_results} shows the accuracy comparison regarding different settings of variable for PEFT methods on math reasoning datasets. The fine-tuning dataset used for this study is Math10K. It can be noted that when the number of virtual tokens in Prefix-Tuning is set to 10, Prefix-Tuning attains an average accuracy of $42.0\%$ on math reasoning datasets. By configuring the bottleneck dimension to 256, Series and Parallel Adapter demonstrate the highest level of performance. However, when the bottleneck size is increased to 512, the accuracy of both Series and Parallel Adapter decreases. The typical setting for LoRA rank is set to 8, but we have discovered that a larger rank can enhance the performance of LoRA. Remarkably, when the rank is increased to 32, LoRA achieves an accuracy of $61.9\%$.

\begin{table*}[t]\centering
\begin{tabular}{llccccccc}
\hline
Model & Variable & MultiArith  &GSM8K &AddSub &AQuA &SingleEq &SVAMP & Average \\\hline


\multirow{4}{*}{Prefix}& vt=10 & 63.2&	24.4&	57.0&	14.2&	55.3&	38.1&	42.0 \\
& vt=20  & 60.3&	22.9&	46.1&	16.1&	51.8&	30.6&	38.0\\
& vt=30 & 51.2&	16.9&	42.3&	15.4&	41.9&	31.1&	33.1\\
& vt=40 & 54.2&	17.8&	49.6&	\textbf{21.7}&	52.0&	33.2&	38.1 \\

\hline

\multirow{4}{*}{Series}& bn=64 & 93.0 &31.6&	80.5&	14.6&	79.9&	44.1&	57.3 \\
& bn=128  & 93.8&	33.4&	81.5&	18.5&	79.9&	48.1&	59.2 \\
& bn=256 & 92.8&	33.3&	80.0&	15.0&	83.5&	\textbf{52.3}&	59.5 \\
& bn=512 & 95.3&	33.7&	77.2&	14.9&	76.6&	42.0&	56.6 \\

\hline

\multirow{4}{*}{Parallel}& bn=64 & 92.7&	36.8&	77.5&	16.9&	81.9&	48.5&	59.1 \\
& bn=128  &95.8	&36.9&	83.0&	17.7&	83.3&	47.8&	60.8 \\
& bn=256 &94.5&	35.3&	86.6&	18.1&	\textbf{86.0}&	49.6&	61.7 \\
& bn=512 &95.0&	31.5&	81.8&	14.6&	81.5&	43.4&	58.0 \\
\hline

\multirow{4}{*}{LoRA}& r=4 & 95.2&	34.7&	78.5&	16.1&	77.8&	45.4&	57.9 \\
& r=8  &\textbf{96.2}&	35.6&	80.5&	15.7&	82.3&	49.6&	60.0 \\
& r=16 & 95.5&	36.2&	82.8&	13.8&	84.4&	50.9&	60.6\\
& r=32 &95.0&	\textbf{37.5}&	\textbf{83.3}&	18.9&	84.4&	52.1&	\textbf{61.9} \\

\hline
\end{tabular}
\caption{The accuracy comparison regarding different settings of variable for PEFT methods on math reasoning datasets. The fine-tuning dataset used for this analysis is Math10K. In this context, "vt" refers to the number of virtual tokens, "bn" denotes the bottleneck size, while "r" is the LoRA rank. The base model employed for this study is LLaMA-7B.
}
\label{tab:configuration_results}

\end{table*}

\subsection{Qualitative Examples}
\label{sec:examples}

We will show examples randomly sampled from math reasoning and commonsense reasoning datasets in this section. 

\begin{table*}[t]\centering
\small
\begin{tabular}{lp{13cm}}\hline
Question &  At the schools book fair Sam bought 13 adventure books and 17 mystery books. If 15 of the books were used, how many new books did he buy?  \\
\hline
ChatGPT & To calculate the number of mosaic tiles needed to cover two-thirds of Mr. Boarden's 36 square foot bathroom, follow these steps:

Find two-thirds of 36: Multiply 36 by 2/3.

(36 * 2/3) = 24 square feet

Determine the number of mosaic tiles needed per square foot: Given that Mr. Boarden needs 24 mosaic tiles per square foot, multiply the result from step 1 by 24.

(24 * 24) = 576 mosaic tiles

Therefore, Mr. Boarden would need 576 mosaic tiles to cover two-thirds of his 36 square foot bathroom.
\\ \hline

LLaMA$_\text{13B}$-Prefix & 1. Find the total number of books Sam bought: 

13 + 17 = 30 books 

2. Subtract the number of used books from the total number of books: 

30 - 15 = 15 new books 

Therefore, Sam bought 15 new books at the book fair. 

Steps: 

1. Total number of books = 13 + 17 = 30 

2. Number of used books = 15 

3. Number of new books = Total number of books - Number of used books 

4. Number of new books = 30 - 15 = 15 

5. Sam bought 15 new books at the book fair. 

Answer: Sam bought 15 new books at the book fair. (15)

\\ \hline

LLaMA$_\text{13B}$-Series & 1. Find the total number of books Sam bought: 13 + 17 = 30

2. Subtract the number of used books from the total number of books: 30 - 15 = 15

3. Therefore, Sam bought 15 new books. 

Answer: 15
\\ \hline

LLaMA$_\text{13B}$-Parallel & To find out how many new books Sam bought, we need to subtract the number of used books from the total number of books he bought.

Total books = 13 + 17 = 30

Used books = 15

New books = Total books - Used books

New books = 30 - 15

New books = 15

Therefore, Sam bought 15 new books.
\\\hline
LLaMA$_\text{13B}$-LoRA &   1. Find the total number of books Sam bought: 13 + 17 = 30

2. Subtract the number of used books from the total number of books: 30 - 15 = 15

3. Therefore, Sam bought 15 new books. 

Answer: 15
\\\hline

\hline
\end{tabular}
\caption{An example randomly sampled from MultiArith. The outputs of ChatGPT and LLaMA-13B with different PEFT methods.}
\end{table*}

\begin{table*}[t]\centering
\small
\begin{tabular}{lp{13cm}}\hline
Question &  Mr Boarden is remodeling his bathroom. For every square foot, he needs 24 mosaic tiles. How many mosaic tiles would Mr Boarden need to cover two thirds of his 36 sq ft bathroom? \\
\hline
ChatGPT & To calculate the number of mosaic tiles needed to cover two-thirds of Mr. Boarden's 36 square foot bathroom, follow these steps: 

1. Find two-thirds of 36: Multiply 36 by 2/3.

\textbf{(36 * 2/3) = 24 square feet}

2.Determine the number of mosaic tiles needed per square foot: Given that Mr. Boarden needs 24 mosaic tiles per square foot, multiply the result from step 1 by 24.

\textbf{(24 * 24) = 576 mosaic tiles}

Therefore, Mr. Boarden would need 576 mosaic tiles to cover two-thirds of his 36 square foot bathroom.
\\ \hline

LLaMA$_\text{13B}$-Prefix & 1 square foot = 4 square inches

1 square inch = 16 square centimeters

So, 1 sq ft = 16 sq cm

Therefore, 36 sq ft = 36 x 16 sq cm = 576 sq cm

To cover two-thirds of the bathroom, Mr Boarden needs to cover 576 sq cm x 2/3 = 384 sq cm

Since 1 sq cm = 1 mosaic tile, Mr Boarden needs 384 sq cm / 1 sq cm = 384 mosaic tiles

Therefore, Mr Boarden needs 384 mosaic tiles to cover two-thirds of his 36 sq ft bathroom. \\ \hline

LLaMA$_\text{13B}$-Series & 1. Find the area of the bathroom: 36 sq ft

2. Find two-thirds of the area: \textbf{2/3 x 36 sq ft = 24 sq ft}

3. Multiply the area by the number of tiles needed per square foot: \textbf{24 sq ft x 24 tiles/sq ft = 576 tiles}

4. Therefore, Mr Boarden would need 576 mosaic tiles to cover two-thirds of his 36 sq ft bathroom. \\ \hline

LLaMA$_\text{13B}$-Parallel &  1. Calculate the total area of the bathroom: 36 sq ft

2. Determine how much area Mr Boarden wants to cover with mosaic tiles: \textbf{2/3 x 36 sq ft = 24 sq ft}

3. Calculate the total number of mosaic tiles needed: \textbf{24 sq ft x 24} mosaic tiles per sq ft = 600 mosaic tiles

Therefore, Mr Boarden would need 600 mosaic tiles to cover two thirds of his 36 sq ft bathroom. The answer in Arabic numerals is 600.
\\\hline
LLaMA$_\text{13B}$-LoRA &  1. Calculate the total area of the bathroom: 36 sq ft

2. Determine how much area needs to be covered: \textbf{2/3 x 36 sq ft = 24 sq ft}

3. Calculate the number of mosaic tiles needed: \textbf{24 sq ft x 24 mosaic tiles per sq ft = 576 mosaic tiles}

Therefore, Mr Boarden would need 576 mosaic tiles to cover two thirds of his 36 sq ft bathroom. The answer in Arabic numerals is 576.  \\\hline

\hline
\end{tabular}
\caption{An example randomly sampled from GSM8K. The outputs of ChatGPT and LLaMA-13B with different PEFT methods.}
\end{table*}

\begin{table*}[t]\centering
\small
\begin{tabular}{lp{13cm}}\hline
Question & There are 7 crayons in the drawer. Mary took 3 crayons out of the drawer. How many crayons are there now?   \\
\hline
ChatGPT & To calculate the number of crayons remaining in the drawer after Mary took some out, follow these steps:

Subtract the number of crayons Mary took from the total number of crayons in the drawer: 7 - 3 = 4 crayons.

Therefore, there are now 4 crayons remaining in the drawer.
\\ \hline

LLaMA$_\text{13B}$-Prefix & 1. Start with the initial number of crayons in the drawer: 7.

2. Subtract the number of crayons Mary took out of the drawer: 3.

3. The final number of crayons in the drawer is 7 - 3 = 4.

Therefore, there are 4 crayons in the drawer now.
\\ \hline

LLaMA$_\text{13B}$-Series & 1. Start with the initial number of crayons in the drawer: 7

2. Subtract the number of crayons Mary took out: 3

3. Perform the subtraction: 7 - 3 = 4

4. Therefore, there are now 4 crayons in the drawer.

Answer: There are now 4 crayons in the drawer.
\\ \hline

LLaMA$_\text{13B}$-Parallel & 1. Start with the total number of crayons in the drawer: 7

2. Subtract the number of crayons Mary took out: 3

3. Perform the subtraction: 7 - 3 = 4

4. The final answer is 4 crayons.

Therefore, there are now 4 crayons in the drawer.
\\\hline
LLaMA$_\text{13B}$-LoRA &    1. Start with the initial number of crayons: 7

2. Subtract the number of crayons Mary took out: 7 - 3 = 4

3. The final answer is 4 crayons.

Therefore, there are now 4 crayons in the drawer.
\\\hline

\hline
\end{tabular}
\caption{An example randomly sampled from AddSub. The outputs of ChatGPT and LLaMA-13B with different PEFT methods.}
\end{table*}

\begin{table*}[t]\centering
\small
\begin{tabular}{lp{13cm}}\hline
Question &  Suppose you want to arrange your English, Hindi, Mathematics, History, Geography and Science books on a shelf. In how many ways can you do it ? Answer Choices: (A) 520 (B) 720 (C) 920 (D) None (E) Cannot be determined  \\
\hline
ChatGPT & To calculate the number of ways you can arrange the English, Hindi, Mathematics, History, Geography, and Science books on a shelf, we can use the concept of permutations. Since order matters when arranging the books, we need to find the number of permutations.

To determine the total number of arrangements, follow these steps:

Count the number of books: There are 6 books in total.

Use the formula for permutations of n objects, where n is the number of books: n! (n factorial).

n! = 6! = 6 x 5 x 4 x 3 x 2 x 1 = 720

Therefore, you can arrange the books on the shelf in 720 different ways.

The correct answer choice is (B) 720.
\\ \hline

LLaMA$_\text{13B}$-Prefix & 1. Let's start by assigning variables to the unknowns in the problem. Let a, b, and c be the digits in M, N, and 396c respectively.

2. We know that M and N are three-digit numbers, so a + b + c = 3.

3. We also know that M - N + 396c = 990, so we can substitute a, b, and c into this equation:

a(1000000 - 100000 + 396c) + b(1000000 - 100000 + 396c) + c(1000000 - 100000 + 396c) = 990

4. Simplifying this equation, we get:

a
\\ \hline

LLaMA$_\text{13B}$-Series & To solve this problem, we need to use the multiplication principle of counting, which states that if there are m ways to do one thing and n ways to do another thing, then there are m x n ways to do both things together.

In this case, there are 6 books to arrange (English, Hindi, Mathematics, History, Geography, and Science). We can arrange these books in 6! ways.

Therefore, the total number of ways to arrange the books on the shelf is 6! = 720.

Hence, the answer is option B) 720.
\\ \hline

LLaMA$_\text{13B}$-Parallel & The total number of books to be arranged on the shelf is 5 (English, Hindi, Mathematics, History, Geography, and Science).

To find the number of ways to arrange these books on the shelf, we can use the formula for permutations of n objects taken r at a time, which is given by:

P(n,r) = n! / (n-r)!

where n is the total number of objects, and r is the number of objects to be arranged at a time.

In this case, n = 5 (the total number of books), and r = 5 (since we want to arrange all 5 books on the shelf).

Substituting these values into the formula, we get:

P(5,5) = 5! / (5-5)!
P(5,5) = 5! / (5-5)!
P(5,5) = 5! / (5-5)!
P(5,5) = 5! / (5-5)!
P(5,5) = 5! / (5-5)!
\\\hline
LLaMA$_\text{13B}$-LoRA &  The total number of ways to arrange 6 books on a shelf is 6! = 6 x 5 x 4 x 3 x 2 x 1 = 720.

Therefore, the answer is option B) 720.  \\\hline

\hline
\end{tabular}
\caption{An example randomly sampled from AQuA. The outputs of ChatGPT and LLaMA-13B with different PEFT methods.}
\end{table*}

\begin{table*}[t]\centering
\small
\begin{tabular}{lp{13cm}}\hline
Question &   Oceanside Bike Rental Shop charges 17 dollars plus 7 dollars an hour for renting a bike. Tom paid 80 dollars to rent a bike. How many hours did he pay to have the bike checked out? \\
\hline
ChatGPT & To calculate the number of hours Tom paid to have the bike checked out, follow these steps:

1. Subtract the base charge from the total amount Tom paid: 80 - 17 = 63 dollars.

2. Divide the remaining amount by the hourly rate to find the number of hours: 63 / 7 = 9 hours.

Therefore, Tom paid to have the bike checked out for 9 hours.

Note: It's important to note that the result represents the whole number of hours. If there were any partial hours, they would not be reflected in the calculation.
\\ \hline

LLaMA$_\text{13B}$-Prefix & 1. Let x be the number of hours Tom paid to rent the bike.

2. The cost of renting the bike for x hours is 17 + 7x dollars.

3. We know that Tom paid 80 dollars, so we can set up an equation:

17 + 7x = 80

4. Solving for x, we divide both sides by 7:

x = 10

5. Therefore, Tom paid 10 hours to rent the bike. 

\\ \hline

LLaMA$_\text{13B}$-Series & 1. Find the total cost of renting the bike: 

\$17 + \$7/hour = \$24/hour 

2. Divide the total cost by the cost per hour to find the number of hours: 

\$80 ÷ \$24/hour = 3.33 hours 

3. Round up to the nearest whole number since you can't rent a bike for a fraction of an hour: 

3 hours 

Therefore, Tom paid to have the bike checked out for 3 hours.
\\ \hline

LLaMA$_\text{13B}$-Parallel & 1. Subtract the initial charge of \$17 from the total amount paid: \$80 - \$17 = \$63

2. Divide the remaining amount by the hourly rate: \$63 ÷ \$7 = 9

3. The answer is 9 hours.

Therefore, Tom paid to have the bike checked out for 9 hours.
\\\hline
LLaMA$_\text{13B}$-LoRA &  1. Let x be the number of hours Tom paid to have the bike checked out.

2. The cost of renting the bike for x hours is 7x dollars.

3. The total cost of renting the bike for x hours is 17 + 7x dollars.

4. We know that Tom paid 80 dollars to rent the bike, so we can set up an equation: 17 + 7x = 80

5. Solving for x, we can subtract 17 from both sides: 7x = 63

6. Dividing both sides by 7, we get x = 9

Therefore, Tom paid to have the bike checked out for 9 hours.  \\\hline

\hline
\end{tabular}
\caption{An example randomly sampled from SingleEq. The outputs of ChatGPT and LLaMA-13B with different PEFT methods.}
\end{table*}

\begin{table*}[t]\centering
\small
\begin{tabular}{lp{13cm}}\hline
Question &  Matthew gave equal numbers of crackers and cakes to his 4 friends. If he had 32 crackers and 98 cakes initially. How many crackers did each person eat?  \\
\hline
ChatGPT & To calculate the number of crackers each person ate, we need to find the number of crackers that each friend received initially and subtract it from the total number of crackers Matthew had.

Here are the steps:

Divide the total number of crackers by the number of friends (4) to find the number of crackers each friend received initially: 32 / 4 = 8 crackers per friend.

Since we want to find the number of crackers each person ate, we need to subtract the initial number of crackers per friend from the total number of crackers: 32 - 8 = 24 crackers.

Therefore, each person ate 24 crackers.
\\ \hline

LLaMA$_\text{13B}$-Prefix & 1. Find the total number of crackers and cakes that Matthew initially had: 32 + 98 = 130

2. Divide the total number of crackers and cakes by the number of friends to find the initial amount given to each friend: 130 ÷ 5 = 26

3. Subtract the initial amount given to each friend from the total number of crackers and cakes to find the remaining amount: 130 - 26 = 104

4. Divide the remaining amount by the number of friends to find the amount each person ate: 104 ÷ 5 = 21

Therefore, each person ate 21 crackers.
\\ \hline

LLaMA$_\text{13B}$-Series & 1. Divide the total number of crackers by the number of friends: 32 ÷ 4 = 8

2. Divide the total number of cakes by the number of friends: 98 ÷ 4 = 24

3. Add the number of crackers and cakes each friend received: 8 + 24 = 32

4. Therefore, each person ate 32 crackers and 24 cakes.

Answer: Each person ate 32 crackers.
\\ \hline

LLaMA$_\text{13B}$-Parallel & 1. Divide the total number of crackers by the number of friends: 32 ÷ 4 = 8

2. Divide the total number of cakes by the number of friends: 98 ÷ 4 = 24

3. Add the number of crackers and cakes each person ate: 8 + 24 = 32

Therefore, each person ate 32 crackers and cakes combined.
\\\hline
LLaMA$_\text{13B}$-LoRA & 1. Divide the total number of crackers by the number of friends: 32 ÷ 4 = 8

2. Divide the total number of cakes by the number of friends: 98 ÷ 4 = 24

3. Add the number of crackers and cakes each person received: 8 + 24 = 32

Therefore, each person ate 32 crackers and cakes in total.   \\\hline

\hline
\end{tabular}
\caption{An example randomly sampled from SVAMP. The outputs of ChatGPT and LLaMA-13B with different PEFT methods.}
\end{table*}

\begin{table*}[t]\centering
\small
\begin{tabular}{lp{13cm}}\hline
Question &  Does ethanol take more energy make that produces?  \\
\hline
ChatGPT & False.
\\ \hline

LLaMA$_\text{13B}$-Prefix & the correct answer is true
\\ \hline

LLaMA$_\text{13B}$-Series & the correct answer is true
\\ \hline

LLaMA$_\text{13B}$-Parallel & false
\\\hline
LLaMA$_\text{13B}$-LoRA &  the correct answer is false  \\\hline

\hline
\end{tabular}
\caption{An example randomly sampled from BoolQ. The outputs of ChatGPT and LLaMA-13B with different PEFT methods.}
\end{table*}

\begin{table*}[t]\centering
\small
\begin{tabular}{lp{13cm}}\hline
Question & How do I ready a guinea pig cage for it's new occupants?

Solution1: Provide the guinea pig with a cage full of a few inches of bedding made of ripped paper strips, you will also need to supply it with a water bottle and a food dish.

Solution2: Provide the guinea pig with a cage full of a few inches of bedding made of ripped jeans material, you will also need to supply it with a water bottle and a food dish.\\
\hline
ChatGPT & Solution1
\\ \hline

LLaMA$_\text{13B}$-Prefix & the correct answer is solution1
\\ \hline

LLaMA$_\text{13B}$-Series & the correct answer is solution2
\\ \hline

LLaMA$_\text{13B}$-Parallel & the correct answer is solution2
\\\hline
LLaMA$_\text{13B}$-LoRA & the correct answer is solution1   \\\hline

\hline
\end{tabular}
\caption{An example randomly sampled from PIQA. The outputs of ChatGPT and LLaMA-13B with different PEFT methods.}
\end{table*}

\begin{table*}[t]\centering
\small
\begin{tabular}{lp{13cm}}\hline
Question & Tracy didn't go home that evening and resisted Riley's attacks. What does Tracy need to do before this?

Answer1: make a new plan Answer2: Go home and see Riley Answer3: Find somewhere to go\\
\hline
ChatGPT & Answer1/make a new plan
\\ \hline

LLaMA$_\text{13B}$-Prefix & the correct answer is answer1
\\ \hline

LLaMA$_\text{13B}$-Series & the correct answer is answer3
\\ \hline

LLaMA$_\text{13B}$-Parallel & the correct answer is answer3
\\\hline
LLaMA$_\text{13B}$-LoRA &  the correct answer is answer3  \\\hline

\hline
\end{tabular}
\caption{An example randomly sampled from SIQA. The outputs of ChatGPT and LLaMA-13B with different PEFT methods.}
\end{table*}

\begin{table*}[t]\centering
\small
\begin{tabular}{lp{13cm}}\hline
Question & Roof shingle removal: A man is sitting on a roof. he

Ending1: is using wrap to wrap a pair of skis. Ending2: is ripping level tiles off. Ending3: is holding a rubik's cube. Ending4: starts pulling up roofing on a roof.   \\
\hline
ChatGPT & Ending4: starts pulling up roofing on a roof.
\\ \hline

LLaMA$_\text{13B}$-Prefix & the correct answer is ending4
\\ \hline

LLaMA$_\text{13B}$-Series & the correct answer is ending4
\\ \hline

LLaMA$_\text{13B}$-Parallel & the correct answer is ending4
\\\hline
LLaMA$_\text{13B}$-LoRA &  the correct answer is ending4  \\\hline

\hline
\end{tabular}
\caption{An example randomly sampled from . The outputs of ChatGPT and LLaMA-13B with different PEFT methods.}
\end{table*}

\begin{table*}[t]\centering
\small
\begin{tabular}{lp{13cm}}\hline
Question &  Sarah was a much better surgeon than Maria so \_ always got the easier cases.

Option1: Sarah Option2: Maria
\\ \hline
ChatGPT & option1
\\ \hline

LLaMA$_\text{13B}$-Prefix & the correct answer is option1
\\ \hline

LLaMA$_\text{13B}$-Series & the correct answer is option1
\\ \hline

LLaMA$_\text{13B}$-Parallel & the correct answer is option1
\\\hline
LLaMA$_\text{13B}$-LoRA &  the correct answer is option1  \\\hline

\hline
\end{tabular}
\caption{An example randomly sampled from WinoGrande. The outputs of ChatGPT and LLaMA-13B with different PEFT methods.}
\end{table*}

\begin{table*}[t]\centering
\small
\begin{tabular}{lp{13cm}}\hline
Question &  Which statement best explains why photosynthesis is the foundation of most food webs?

Answer1: Sunlight is the source of energy for nearly all ecosystems. Answer2: Most ecosystems are found on land instead of in water. Answer3: Carbon dioxide is more available than other gases. Answer4: The producers in all ecosystems are plants. \\
\hline
ChatGPT & 
\\ \hline

LLaMA$_\text{13B}$-Prefix & Answer1. Sunlight is the source of energy for nearly all ecosystems.
\\ \hline

LLaMA$_\text{13B}$-Series & the correct answer is answer1
\\ \hline

LLaMA$_\text{13B}$-Parallel & the correct answer is answer4
\\\hline
LLaMA$_\text{13B}$-LoRA &  the correct answer is answer1  \\\hline

\hline
\end{tabular}
\caption{An example randomly sampled from ARC-e. The outputs of ChatGPT and LLaMA-13B with different PEFT methods.}
\end{table*}

\begin{table*}[t]\centering
\small
\begin{tabular}{lp{13cm}}\hline
Question & An astronomer observes that a planet rotates faster after a meteorite impact. Which is the most likely effect of this increase in rotation?

Answer1: Planetary density will decrease. Answer2: Planetary years will become longer. Answer3: Planetary days will become shorter. Answer4: Planetary gravity will become stronger.   \\
\hline
ChatGPT & Answer3
\\ \hline

LLaMA$_\text{13B}$-Prefix & the correct answer is answer3
\\ \hline

LLaMA$_\text{13B}$-Series & the correct answer is answer2
\\ \hline

LLaMA$_\text{13B}$-Parallel & the correct answer is answer3
\\\hline
LLaMA$_\text{13B}$-LoRA &  the correct answer is answer3  \\\hline

\hline
\end{tabular}
\caption{An example randomly sampled from ARC-c. The outputs of ChatGPT and LLaMA-13B with different PEFT methods.}
\end{table*}

\begin{table*}[t]\centering
\small
\begin{tabular}{lp{13cm}}\hline
Question & A person wants to start saving money so that they can afford a nice vacation at the end of the year. After looking over their budget and expenses, they decide the best way to save money is to

Answer1: make more phone calls Answer2: quit eating lunch out Answer3: buy less with monopoly money Answer4: have lunch with friends   \\
\hline
ChatGPT & Answer2: quit eating lunch out.
\\ \hline

LLaMA$_\text{13B}$-Prefix & the correct answer is answer2
\\ \hline

LLaMA$_\text{13B}$-Series & the correct answer is answer2
\\ \hline

LLaMA$_\text{13B}$-Parallel & the correct answer is answer2
\\\hline
LLaMA$_\text{13B}$-LoRA &  the correct answer is answer2  \\\hline

\hline
\end{tabular}
\caption{An example randomly sampled from OBQA. The outputs of ChatGPT and LLaMA-13B with different PEFT methods.}
\end{table*}

\end{document}